\documentclass[lettersize,journal]{IEEEtran}
\usepackage{amsmath,amsfonts}
\usepackage{algorithmic}
\usepackage{algorithm}
\usepackage{array}
\usepackage[caption=false,font=normalsize,labelfont=sf,textfont=sf]{subfig}
\usepackage{textcomp}
\usepackage{stfloats}
\usepackage{url}
\usepackage{verbatim}
\usepackage{graphicx}
\usepackage{cite}
\usepackage{amssymb}
\usepackage{booktabs}
\usepackage{multirow}
\usepackage{multicol}
\usepackage{amsmath}
\usepackage{amsthm}
\usepackage{bm}
\usepackage{epstopdf}
\usepackage{hyperref}
\usepackage{subcaption}
\newtheorem{remark}{Remark}
\begin{document}

\title{AS-GCL: Asymmetric Spectral Augmentation on Graph Contrastive Learning}

\author{Ruyue Liu, Rong Yin\textsuperscript{*}, Yong Liu, Xiaoshuai Hao, Haichao Shi, Can Ma, and Weiping Wang
\thanks{Ruyue Liu, Rong Yin, Haichao Shi, Can Ma, and Weiping Wang are with the Institute of Information Engineering, Chinese Academy of Sciences, Beijing 100085, China. Ruyue Liu is with School of Cyberspace Security, University of Chinese Academy of Sciences, Beijing 100049, China.  
E-mail: \{liuruyue, yinrong, shihaichao, macan, wangweiping\}@iie.ac.cn.}

\thanks{Yong Liu is with Renmin University of China, Beijing 100872, China.
E-mail: liuyonggsai@ruc.edu.cn.}

\thanks{Xiaoshuai Hao is with Beijing Academy of Artificial Intelligence, Beijing 100028, China.
E-mail: xshao@baai.ac.cn.}

\thanks{Rong Yin is the corresponding author.}}


\maketitle

\begin{abstract}
Graph Contrastive Learning (GCL) has emerged as the foremost approach for self-supervised learning on graph-structured data. GCL reduces reliance on labeled data by learning robust representations from various augmented views. However, existing GCL methods typically depend on consistent stochastic augmentations, which overlook their impact on the intrinsic structure of the spectral domain, thereby limiting the model's ability to generalize effectively.
To address these limitations, we propose a novel paradigm called AS-GCL that incorporates asymmetric spectral augmentation for graph contrastive learning. A typical GCL framework consists of three key components: graph data augmentation, view encoding, and contrastive loss. Our method introduces significant enhancements to each of these components. Specifically, for data augmentation, we apply spectral-based augmentation to minimize spectral variations, strengthen structural invariance, and reduce noise. With respect to encoding, we employ parameter-sharing encoders with distinct diffusion operators to generate diverse, noise-resistant graph views. For contrastive loss, we introduce an upper-bound loss function that promotes generalization by maintaining a balanced distribution of intra- and inter-class distance.
To our knowledge, we are the first to encode augmentation views of the spectral domain using asymmetric encoders. Extensive experiments on eight benchmark datasets across various node-level tasks demonstrate the advantages of the proposed method.
\end{abstract}


\begin{figure*}[!t]
\centering
\includegraphics[width=0.87\textwidth]{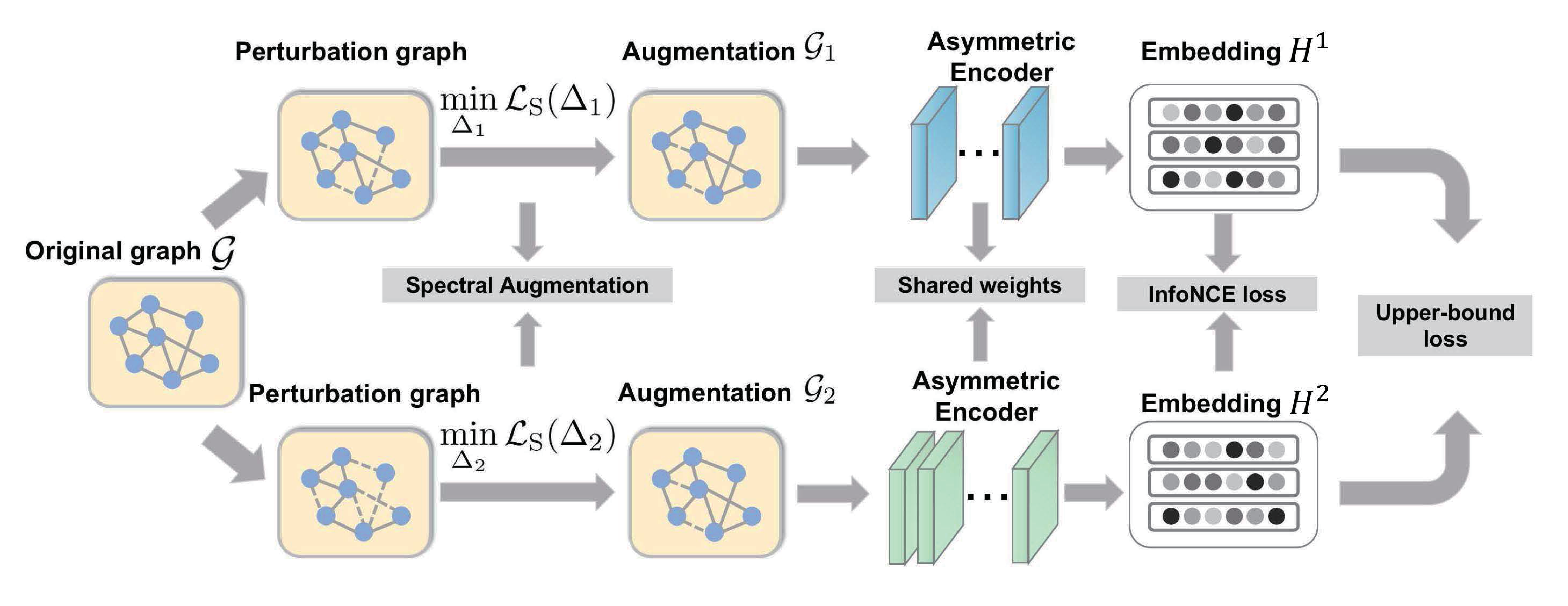}
\caption{The general framework of the proposed AS-GCL method. First, topology augmentation is optimized by generating paired augmented views \( \mathcal{G}_1 \) and \( \mathcal{G}_2 \) through spectral variation minimization. Then, encoders with shared parameters but different diffusion operators are used to generate representations for different views (\( \boldsymbol{H}^1 \), and \( \boldsymbol{H}^2 \)). Finally, a contrastive loss function is introduced to reduce intraclass variation and increase interclass contrast.}
\label{fig1}
\end{figure*}

\section{Introduction}
\IEEEPARstart{G}{raph} data play a significant role in multimedia applications, enabling cross-modal content analysis and generation. Graph neural networks (GNNs)  have been widely adopted across diverse domains such as computer vision (CV) \cite{9712249}, natural language processing (NLP) \cite{9354543}, and information retrieval (IR) \cite{10292891}. Notably, Graph Convolutional Networks (GCNs) \cite{kipf2016semi} and Graph Attention Networks (GATs) \cite{velivckovic2017graph} are specifically tailored to semi-supervised node classification tasks involving labeled nodes. Labeled data are particularly crucial for enhancing the performance of these GNN models.

Contrastive learning (CL) learns representations from supervision signals derived from data without relying on human-provided labels. CL has achieved great success in many domains, including computer vision \cite{10243145}, natural language processing \cite{rethmeier2023primer}, and signal processing \cite{10286310}. Specifically, contrastive methods aim to build effective representations by joining semantically similar (positive) pairs and separating dissimilar (negative) pairs \cite{wang2022uncovering,wang2022augmentation}. Inspired by the success of contrastive methods in computer vision, these methods have recently been adopted for graphs \cite{10286310,zhang2022costa,LIU2024110274}.

While graph contrastive learning methods are effective on graph-related tasks, they often overlook a fundamental distinction between images and graphs. In the context of images, augmentation is well-defined and typically involves operations like cropping and rotation. However, in graphs, the representation of augmentation can become arbitrary. For example, the core semantics in images remain relatively stable even after they undergo random transformations \cite{gong2023ma}. In contrast, when the edges or nodes of a graph are perturbed, it is difficult to determine whether the augmented graph maintains a positive correlation with the original graph. This is mainly because graphs contain both semantic and structural information.

Due to the arbitrary behavior of augmentation in graphs, the quality of learned graph representations in augmentation-based contrastive methods highly depends on selecting an appropriate augmentation scheme \cite{lee2022augmentation}. Graph augmentation methods can be categorized into four main groups: subgraph extraction \cite{sun2019infograph,xia2022substructure}, adaptive augmentation \cite{jing2022graph,wang2021adaptive}, random augmentation \cite{hassani2020contrastive,wang2022uncovering,liu2024aswt}, and other methods \cite{zhang2022costa,xia2022simgrace}. Subgraph extraction involves partitioning each graph into multiple subgraphs, with subgraph selection determined by the specific requirements of different models. However, this category of methods incurs significant computational overhead that is often considered prohibitive. Adaptive augmentation leverages principles rooted in task labeling and information bottlenecks to identify optimal views to enhance learning, although these methods usually entail intricate model architectures. In random augmentation, contrastive views are constructed by applying random perturbations, such as discarding nodes or edges. While straightforward and readily implementable, random augmentation overlooks the nuanced impact of distinct edge perturbations on underlying graph structure, thus failing to benefit from the relational inductive bias of graph-structured data.

To address the severe structural damage caused by consistent random augmentation processes, we propose AS-GCL, a contrastive learning method utilizing spectral augmentation. To our knowledge, AS-GCL represents a pioneering effort in encoding spectral augmented views through an asymmetric encoder. The proposed method consists of three main components. The first component employs minimal spectral variation to guide topology augmentation, resulting in augmented views with minimal structural differences. In the second phase, we improve the GCN encoder by using an asymmetric encoder to generate different view representations. The final component achieves model optimization by maximizing the mutual information of the various view representations. We introduce a multiple loss function based on upper-bound loss, which prevents positive embeddings from deviating significantly from anchor embeddings, thereby minimizing intraclass differences. The main contributions of this paper are as follows.

\begin{itemize}
\item We propose the AS-GCL framework, employing a novel asymmetric spectral augmentation strategy that minimizes spectral variations and generates diverse, noise-resilient graph views through parameter-sharing encoders with different diffusion operators.

\item This method introduces spectral-based augmentation and an upper-bound loss function, ensuring improved structural invariance, reduced noise, and improved generalizability by maintaining balanced intraclass and interclass distance.

\item Extensive experiments on node classification and clustering tasks demonstrate that AS-GCL consistently outperforms state-of-the-art methods and shows strong robustness against adversarial attacks on graph structures.
\end{itemize}

The paper is organized as follows: Section \ref{Related Work} provides an overview of related work on graph contrastive learning. Section \ref{Preliminaries} introduces the notation and preliminaries used in the paper. Section \ref{Method} presents the details of the proposed method. Section \ref{Experimental Evaluation} describes a comprehensive set of experiments conducted to evaluate the effectiveness of the proposed method. Finally, Section \ref{conclusion} summarizes the contributions of the paper and describes future research directions.

\section{Related Work}
\label{Related Work}
\subsection{Graph Contrastive Learning}
Recently, due to the significant success of contrastive learning methods for images, CL techniques have increasingly been applied to graphs. DGI \cite{velivckovic2018deep} is a notable pioneering work inspired by Deep InfoMax \cite{hjelm2018learning}. The objective of DGI is to acquire node representations by maximizing the mutual information of local graph blocks. GMI \cite{peng2020graph} and HDMI \cite{jing2021hdmi} further enhanced DGI by incorporating mutual information about edges and node attributes. Inspired by SimCLR \cite{chen2020simple}, GRACE \cite{zhu2020deep}, which generates two augmented graph views by randomly perturbing nodes and edges, learns node representations by aligning representations of the same nodes across two augmented graphs while maintaining separation of the other nodes. Despite these advancements, a major criticism of graph contrastive learning (GCL) methods is so-called semantic drift, where the augmentation process can lead to distorted or inconsistent node representations, thereby reducing the quality of learned embeddings \cite{sun2021mocl}. Additionally, most existing GCL frameworks use two view encoders with identical architectures, which may limit their ability to learn diverse features from augmented graph views \cite{sun2021mocl}. In this study, we propose a novel paradigm for GCL. Unlike existing methods, the proposed AS-GCL method employs asymmetric view encoders with identical parameters but a different number of propagation layers. This design enables the encoders to capture information from both long-range and short-range connections in the graph, resulting in richer and more diverse representations. Structural asymmetry helps enhance the quality of the learned representations and opens new avenues for advancing contrastive learning on graph data.

\subsection{Spectrum-based Augmentation}

Most existing augmentation methods focus on the spatial domain, with relatively few studies exploring the spectral domain. Spectral-based methods can be broadly categorized as either utilizing the graph Laplacian spectrum or the feature spectrum. Methods leveraging the graph Laplacian spectrum focus primarily on capturing global graph structure. For example, SpCo \cite{liu2022revisiting} aims to capture the low-frequency information common to both augmentations by adjusting the balance of high- and low-frequency components. Similarly, Amur Ghose et al. \cite{ghose2023spectral} proposed a method that enhances representations by rearranging various frequency components within graphs. GCIL \cite{mo2024graph} adopts a causal perspective by considering low-frequency components as causal variables and high-frequency components as noncausal variables. GCIL achieves graph augmentation by perturbing noncausal information and retaining causal information.
Feature spectrum-based methods emphasize the local features of nodes or edges. For example, COSTA \cite{zhang2022costa} decomposes hidden space representations and optimizes the alignment process by adding uniform noise (perturbation) to singular values. SFA \cite{zhang2023spectral} balances the contribution of each component within the feature spectrum to achieve an optimal representation. These methods typically rely on spectral filters to balance low- and high-frequency components, which requires careful tuning based on specific datasets and tasks.
In contrast, our method focuses on minimizing Laplacian spectral variations to effectively mitigate the semantic drift caused by graph data augmentation. Unlike previous methods, our method does not require fine-tuning of frequency components for each dataset; rather, it automatically learns the structure with minimal variance. By reducing Laplacian spectral variation, our method ensures a more consistent global representation, leading to improved generalizability and robustness across different tasks.

\section{Preliminaries}
\label{Preliminaries}
Let \(\mathcal{G} = (\mathcal{V}, \mathcal{E})\) represent an undirected graph, where \(\mathcal{V} = \{v_{1}, \ldots, v_{N}\}\) denotes the set of \(N\) nodes, and \(\mathcal{E} \subseteq \mathcal{V} \times \mathcal{V}\) denotes the set of edges. The existence of edges is represented by an adjacency matrix \(\boldsymbol{A} \in \{0, 1\}^{N \times N}\), where \(\boldsymbol{A}_{ij} \in \{0, 1\}\) signifies the relationship between nodes \(v_i\) and \(v_j\) in \(\mathcal{V}\). Additionally, graph \(\mathcal{G}\) is associated with a node feature matrix \(\boldsymbol{X} = [x_1, x_2, \ldots, x_N] \in \mathbb{R}^{N \times d}\), where \(x_i\) represents a \(d\)-dimensional feature vector of node \(v_i \in \mathcal{V}\). The node degree matrix is denoted as \(\boldsymbol{D} = \operatorname{diag}(d_1, \ldots, d_N)\), where \(d_i = \sum_{v_j \in \mathcal{V}} \boldsymbol{A}_{ij}\) represents the degree of node \(v_i \in \mathcal{V}\). The objective is to learn (in an unsupervised manner) a GNN encoder \(f_{\theta}: (\boldsymbol{X}, \boldsymbol{A}) \to \boldsymbol{H} \in \mathbb{R}^{N \times m}\), which takes node features and graph structure as inputs and generates low-dimensional node representations, where \(m\) is significantly smaller than \(d\).

Let \(\boldsymbol{L} = \boldsymbol{D} - \boldsymbol{A}\) represent the unnormalized graph Laplacian of \(\mathcal{G}\). If we denote the symmetric normalized adjacency matrix as \(\hat{\boldsymbol{A}} = \boldsymbol{D}^{-\frac{1}{2}} \boldsymbol{A} \boldsymbol{D}^{-\frac{1}{2}}\), then the symmetric normalized graph Laplacian is \(\hat{\boldsymbol{L}} = \boldsymbol{I} - \hat{\boldsymbol{A}} = \boldsymbol{D}^{-\frac{1}{2}} \boldsymbol{L} \boldsymbol{D}^{-\frac{1}{2}}\). Since \(\hat{\boldsymbol{L}}\) is symmetric and normalized, its eigen-decomposition is \(\boldsymbol{U} \boldsymbol{\Lambda} \boldsymbol{U}^{\top}\), where \(\boldsymbol{\Lambda} = \operatorname{diag}(\lambda_{1}, \ldots, \lambda_{N})\) contains the eigenvalues of \(\hat{\boldsymbol{L}}\), and \(\boldsymbol{U} = [\boldsymbol{u}_{1}, \ldots, \boldsymbol{u}_{N}] \in \mathbb{R}^{N \times N}\) contains the eigenvectors of \(\hat{\boldsymbol{L}}\).

\section{Proposed Method}
\label{Method}
We propose a simple yet effective graph contrastive learning framework, AS-GCL, for generating more specific graph representations by leveraging the relational inductive bias of graph-structured data. The proposed method includes three key improvements: an augmentation scheme, an encoder, and a contrastive loss function. In AS-GCL's augmentation scheme, the graph spectrum is utilized to capture structural properties, and the invariance of the graph spectrum is employed to represent structural invariance. With respect to the encoder, AS-GCL uses asymmetric encoders with a different number of diffusion layers to generate view embeddings. For contrastive loss, we introduce an upper-bound loss component to constrain the distance between positive and negative pairs.

\subsection{Spectral Augmentation}
\label{sa}
The objective of our spectral augmentation scheme is to reduce dependence on unstable components and effectively mitigate the noise associated with data augmentation. Specifically, we define an edge perturbation-based topology augmentation scheme determined by a Bernoulli probability matrix. The augmentation principle is formulated as an optimization problem based on this topological augmentation scheme to minimize the spectral variation of the augmented graphs.

\textbf{Topology Augmentation.} To enhance graph topology diversity and improve the robustness of graph representation learning, we propose an edge perturbation-based topology augmentation scheme. This scheme introduces a Bernoulli probability matrix to control the probability of edge flipping, enabling the generation of diverse graph structures to enhance generalizability. Specifically, for a given adjacency matrix \(\boldsymbol{A}\), we define topological augmentation matrix \(\tilde{\boldsymbol{A}}\) as the result of sampling from a Bernoulli distribution \(\mathcal{B}(\Delta)\), where \(\Delta \in [0, 1]^{N \times N}\) is the parameter matrix that controls the probability of edge flipping. The element \(\Delta_{ij}\) represents the probability of flipping the edge between nodes \(i\) and \(j\), which can be flexibly adjusted based on prior knowledge or experimental experience.

Our choice to use a Bernoulli probability matrix to control edge perturbation is motivated by its flexibility and adaptability. Specifically, by adjusting the values in \(\Delta_{ij}\), we can control the intensity of the perturbation. This capability enables us to adjust the intensity of perturbations according to dataset requirements or specific experimental conditions. For example, 
using a lower perturbation probability helps maintain global structural graph integrity
, ensuring that fundamental relationships between nodes are preserved. Conversely, a higher perturbation probability introduces more significant local changes, which can generate more diverse graph views. This flexibility provides balance during augmentation, enabling the model to retain the global characteristics of the original graph while introducing random perturbations to enhance robustness and generalizability across different views. Operationally, we first sample the edge perturbation matrix \(\boldsymbol{M} \in \{0, 1\}^{N \times N}\), where \(\boldsymbol{M}_{ij} \sim \mathcal{B}(\Delta_{ij})\) determines whether to flip the edge between nodes \(i\) and \(j\). If \(\boldsymbol{M}_{ij} = 1\), the edge is flipped; otherwise, it remains unchanged. The topological augmentation matrix is defined as follows:
\begin{equation}
\begin{split}
\boldsymbol{C} &= \Bar{\boldsymbol{A}} - \boldsymbol{A},\\
\tilde{\boldsymbol{A}} &= \boldsymbol{A} + \boldsymbol{C} \circ \boldsymbol{M},
\end{split}
\label{eq2}
\end{equation}
where \(\Bar{\boldsymbol{A}} = \mathbf{1}_N \mathbf{1}_N^{\top} - \boldsymbol{A}\) is the complement matrix of \(\boldsymbol{A}\), \(\mathbf{1}_N\) is an \(N\)-dimensional column vector of ones, and \(\mathbf{1}_N \mathbf{1}_N^{\top}\) represents its outer product. To ensure effective edge perturbation, we introduce the matrix \(\boldsymbol{C}\), where \(\boldsymbol{C}_{ij}\) indicates whether an edge operation is valid between nodes \(i\) and \(j\). If \(\boldsymbol{C}_{ij} = 1\), edge addition is allowed, whereas \(\boldsymbol{C}_{ij} = -1\) allows edge deletion. This method enables the model to adapt to various structural changes during self-supervised learning, ultimately improving its robustness.

\textbf{Minimization of spectral variations.} We optimize \(\Delta\) guided by the graph spectral domain instead of using a fixed value as employed in uniform perturbation. We aim to find the matrix \(\Delta\) that minimizes the expected discrepancy in the graph spectra between the original graph and the graph augmented using the topology defined by \(\Delta\).

To find the desired perturbation matrix \(\Delta\), we formulate the following optimization problem:
\begin{equation}
\begin{split}
\min \mathcal{L}_{S} &= \min_{\Delta \in \mathcal{S}} \| \operatorname{eig}( \text{Lap}(\boldsymbol{A} + \boldsymbol{C} \circ \Delta) ) - \operatorname{eig}( \text{Lap}(\boldsymbol{A}) ) \|_2^2,
\end{split}
\label{eq3}
\end{equation}
where \(\mathcal{S} = \{s \mid s \in [0, 1]^{N \times N}, \|s\|_0 \leq \epsilon \times N \times N\}\), with \(\epsilon\) representing the strength of graph perturbation. \(\text{Lap}(\boldsymbol{A})\) represents the normalized Laplacian matrix of \(\boldsymbol{A}\), and \(\operatorname{eig}(\cdot)\) computes the eigenvalue vector.

We apply two perturbation ratios of varying sizes to enhance the graph topology. By solving Eq. \eqref{eq3}, we derive optimal pairwise Bernoulli probability matrices, \( \Delta_1 \) and \( \Delta_2 \), which produce two augmented views, 
\( \mathcal{G}_1 \) and \( \mathcal{G}_2 \), 
to maintain structural invariance while minimizing deviation from the original graph. This enables the exploration and selection of a strategy closely aligned with the original graph's structural properties. Sampling the augmented views using the optimal \( \Delta \) matrix effectively improves the stability and robustness of the graph representation while preserving key topological features.

To update the value of \(\Delta\)\footnote{Given that $\Delta_1$ and $\Delta_2$ are updated in an identical manner, we simplify the notation by omitting the subscripts and adopting $\Delta$ for consistency.}, we use the following equation:
\begin{equation}
\begin{split}
\Delta^{(t)} &= \Delta^{(t-1)} - \eta_t \nabla \mathcal{L}_S(\Delta^{(t-1)}),
\end{split}
\label{eq4}
\end{equation}
where \(\eta_t > 0\) represents the learning rate at step \(t\). The gradient \(\nabla \mathcal{L}_S(\Delta^{(t-1)})\) can be computed using the chain rule. For a real symmetric matrix \(\boldsymbol{L}\), the derivative of its \(k\)-th eigenvalue \(\lambda_k\) is given by \(\partial \lambda_{k} / \partial \boldsymbol{L} = \mathbf{u}_{k} \mathbf{u}_{k}^{\top}\), where \(\mathbf{u}_{k}\) is the corresponding eigenvector.

Notably, the derivative computation requires distinct eigenvalues, which may not hold for graphs that exhibit self-isomorphism. To address this limitation, we introduce a small noise term to the adjacency matrix, represented as \(\varepsilon \times (\boldsymbol{E} + \boldsymbol{E}^{\top}) / 2\), where each entry in \(\boldsymbol{E}\) is sampled from the uniform distribution \(U(0,1)\), and \(\varepsilon\) is a small constant. The inclusion of the term \((\boldsymbol{E} + \boldsymbol{E}^{\top}) / 2\) ensures that the perturbed adjacency matrix maintains its symmetry, which is particularly important for undirected graphs. By averaging \(\boldsymbol{E}\) with its transpose, we ensure that the resulting perturbed adjacency matrix retains symmetric properties.

\begin{remark}
Directly capturing the structural invariance of graph contrastive learning methods is a complex challenge requiring the simultaneous consideration of various structural attributes. Fortunately, the graph spectrum provides a comprehensive summary of several structural properties, including clustering, connectivity, and diameter \cite{spielman2007spectral,dorfler2018electrical}. Empirically, relatively significant changes in the spectrum correspond to edge flips between nodes from different clusters, indicating more substantial structural alterations. Therefore, the perturbation spectrum can be used to control changes in these structural properties.
By leveraging the graph spectrum to capture structural properties, our goal is to maintain spectral invariance as a proxy for structural invariance. Through graph spectral-guided optimization, we aim to enhance the spectral invariance of augmented graph representations, thereby improving both the accuracy and robustness of the models.
\end{remark}

\subsection{Asymmetric Encoders} In contrastive learning (CL), the learned representation includes both task-relevant information and task-irrelevant noise. CL methods aim to extract consistent information across different views while filtering out task-irrelevant noise specific to any single view. This requires a careful balance: the two views should not be too distant, as this reduces the amount of task-relevant information, nor should they be too close, as this can introduce excessive task-irrelevant noise.
In the proposed spectral augmentation method (Section \ref{sa}), we maintain the structural invariance of graph augmentations, thereby enhancing the consistency between views. However, this method may also increase the proximity between views, potentially leading to reduced differentiation. To address this issue, we employ asymmetric encoders with shared weight parameters but a different number of diffusion layers. This design helps manage the trade-off between view consistency and proximity, improving the overall effectiveness of representation learning.

We decompose each layer of the graph convolutional network (GCN) into two distinct operators, a diffusion operator and a transformation operator, denoted as \(f \) and \(g \), respectively. These operators are defined as follows:
\begin{equation}
\begin{split}
f(\boldsymbol{H}) &=  \boldsymbol{D}^{-\frac{1}{2}} \boldsymbol{A} \boldsymbol{D}^{-\frac{1}{2}} \boldsymbol{H},\\
g(\boldsymbol{H}) &= \sigma(\boldsymbol{H}\boldsymbol{W}),
\end{split}
\label{eq5}
\end{equation}
where \( \boldsymbol{H} \) represents the node embeddings, \( \boldsymbol{D}^{-\frac{1}{2}} \boldsymbol{A} \boldsymbol{D}^{-\frac{1}{2}} \) is the graph filter matrix that performs diffusion, \( \boldsymbol{W} \) is the weight matrix for the linear transformation, and \( \sigma \) denotes a nonlinear activation function. This decomposition allows us to first capture the structural information of the graph through the diffusion operator and then transform the features with a linear transformation followed by a nonlinear activation function.

We express the \(l\)-layer GCN as a combination of multiple diffusion and transformation operators. Specifically, the \(l\)-layer GCN can be written as:
\begin{equation}
\begin{split}
GCN(\boldsymbol{X}) = g^{(l)} \circ f \circ g^{(l-1)} \circ f \circ \cdots \circ g^{(1)} \circ f(\boldsymbol{X}),
\end{split}
\label{eq6}
\end{equation}
where \( \circ \) denotes the composition of two operators and \( g^{(l)} \) represents the transformation operator applied in the \(l\)-th layer of the GCN. This expression shows that the GCN is formed by applying the diffusion operator \(f \) followed by the transformation operators \(g^{(i)} \) for each layer iteratively.

The proposed asymmetric encoder shares weight parameters but employs a different number of diffusion layers. This can be mathematically formalized as follows:
\begin{equation}
\begin{split}
\boldsymbol{H}^1 &= g^{(i)} \circ f \circ \cdots \circ g^{(1)} \circ f(\boldsymbol{X}),\\
\boldsymbol{H}^2 &= f^{[k]} \circ g^{(i)} \circ f \circ \cdots \circ g^{(1)} \circ f(\boldsymbol{X}),
\end{split}
\label{eq7}
\end{equation}
where \( \boldsymbol{H}^1 \) and \( \boldsymbol{H}^2 \) represent the node embeddings generated by two different encoders. Here, \(f^{[k]} \) denotes the composition of \(k \) diffusion operators \(f \). Both encoders share \(i \) transformation layers, which include weight parameters and activation functions. However, the two encoders differ in the number of diffusion layers applied: the first encoder uses \(i \) diffusion layers, whereas the second encoder uses \(i + k \) diffusion layers. These diffusion layers propagate the hidden representations from one layer to the next via the adjacency matrix. Note that the exact number of diffusion layers may vary depending on the design choices and specific requirements of the asymmetric encoder.

\begin{remark}
By employing a varying number of diffusion layers, the two views generated by the asymmetric encoder 
are effectively pushed away from each other while ensuring that they are not overly distant.
 This design maintains balance between the views, preventing them from becoming too dissimilar. Moreover, the weight parameters shared between the encoders ensure that the views remain within an appropriate proximity range. This method enhances view diversity while minimizing noise in the graph convolutional network, thereby improving the robustness and accuracy of graph representation learning.
\end{remark}

\subsection{Optimization Objective}
\textbf{InfoNCE loss.}
We employ a contrastive objective to evaluate node representations obtained from two different graph augmentations. For a given node \(v_a\), the representations \(\boldsymbol{H}^1_a\) and \(\boldsymbol{H}^2_a\) from the respective graph augmentations form a positive pair. Conversely, the representations of other nodes in the two augmentations are considered negative pairs. The paired objective for each positive pair \((\boldsymbol{H}^1_a, \boldsymbol{H}^2_a)\) is defined as follows:
\begin{equation}
\begin{aligned}
\mathcal{L}(\boldsymbol{H}^1_a,\boldsymbol{H}^2_a) = & \\
-\log &\frac{e^{\theta\left(\boldsymbol{H}^1_a, \boldsymbol{H}^2_a\right) }}{e^{\theta\left(\boldsymbol{H}^1_a, \boldsymbol{H}^2_a\right) }+\sum_{b \neq a} \left(e^{\theta\left(\boldsymbol{H}^1_a, \boldsymbol{H}^2_b\right) }+e^{\theta\left(\boldsymbol{H}^1_a, \boldsymbol{H}^1_b\right)}\right)},
\end{aligned}
\label{eq8}
\end{equation}
where \(\theta(\cdot)\) denotes the cosine similarity function. The term \(e^{\theta(\boldsymbol{H}^1_a, \boldsymbol{H}^2_a)}\) in the numerator represents the similarity between the same node \(a\) in two different views, essentially serving as a measure of similarity between positive samples. In the denominator, \(e^{\theta(\boldsymbol{H}^1_a, \boldsymbol{H}^2_a)}\) represents the similarity for the positive pair, whereas the second term sums over all the negative sample pairs. Specifically, \(e^{\theta(\boldsymbol{H}^1_a, \boldsymbol{H}^2_b)}\) represents negative pairs across views, i.e., the similarity between node \(a\) in the first view and node \(b\) in the second view. In addition, \(e^{\theta(\boldsymbol{H}^1_a, \boldsymbol{H}^1_b)}\) indicates the similarity between nodes \(a\) and \(b\) within the same view. This formulation ensures that the negative sample comparisons encompass a broad range of potential combinations, promoting robust contrastive learning.

The overall objective is to maximize the average similarity of all positive pairs, which is expressed as:
\begin{equation}
\begin{split}
\mathcal{L}_{InfoNCE} = \frac{1}{2N} \sum_{a=1}^{N} \left[\mathcal{L}\left(\boldsymbol{H}^1_a, \boldsymbol{H}^2_a\right) + \mathcal{L}\left(\boldsymbol{H}^2_a, \boldsymbol{H}^1_a\right)\right].
\end{split}
\label{eq9}
\end{equation}

\textbf{Upper-bound loss.}
We propose an upper-bound loss function based on triplet state loss to 
address 
intraclass variation and enhance interclass variation. Specifically, the triplet loss for each positive pair \((\boldsymbol{H}^1_a, \boldsymbol{H}^2_a)\) is formulated as:
\begin{equation}
\mathcal{L}_{lower}= \sum_{a=1}^{N} \max\left(0, \left\| \boldsymbol{H}^1_a - \boldsymbol{H}^2_a \right\|_{2} - \left\| \boldsymbol{H}^1_a - \boldsymbol{H}^-_a \right\|_{2} + \alpha \right),
\label{eq11}
\end{equation}
where \(\boldsymbol{H}^-_a\) denotes the negative pair of \(\boldsymbol{H}^1_a\) and \(\alpha\) is a hyperparameter that controls the minimum expected distance between the positive and negative pairs.

While this lower-bound loss \(\mathcal{L}_{lower}\) effectively separates negative pairs, it can inadvertently increase the distance between positive pairs. To address this issue, we introduce an upper-bound loss to constrain the distance between negative and positive pairs. This constraint helps reduce intraclass variation and maintain the proximity of positive samples. The upper-bound loss is defined as:
\begin{equation}
\mathcal{L}_{upper}= \sum_{a=1}^{N} \max\left(0, \left\| \boldsymbol{H}^1_a - \boldsymbol{H}^-_a \right\|_{2} - \left\| \boldsymbol{H}^1_a - \boldsymbol{H}^2_a \right\|_{2} - \beta \right),
\label{eq12}
\end{equation}
where \(\beta\) is a hyperparameter that defines the acceptable range of distances between positive and negative pairs.

Combining InfoNCE, lower-bound, and upper-bound loss, the overall loss function for the proposed method AS-GCL is expressed as:
\begin{equation}
\mathcal{L} = \mathcal{L}_{InfoNCE} + \mathcal{L}_{lower} + \mathcal{L}_{upper}.
\label{eq13}
\end{equation}

This comprehensive loss function ensures effective contrastive learning by balancing the distance between positive and negative pairs, thus enhancing the model's ability to generate robust and accurate graph representations.

\begin{remark}
InfoNCE loss aims to minimize the distance between positive pairs and maximize the distance between negative pairs, which fosters the learning of effective graph representations. However, it does not directly address the issue of the large intraclass distance that can arise if the distance between an anchor and its positive embedding becomes too large. When this happens, the distance between the anchor and negative embeddings can become excessively large, potentially leading to infinite distances. Such large intraclass variation can negatively impact the generalization performance of the model. To address this limitation, AS-GCL incorporates both upper-bound and lower-bound loss. Upper bound loss ensures that the distance between the positive and anchor embeddings remains finite, effectively controlling intraclass variation. This constraint prevents the distance between positive pairs from becoming excessively large, maintaining a reasonable range for the distances between negative and positive pairs. By imposing this balance, the upper-bound loss helps to avoid extreme distances that could impair model performance. In summary, while InfoNCE loss focuses on differentiating between positive and negative pairs, the upper-bound loss in AS-GCL ensures that positive embeddings remain reasonably close to the anchor. This combined method promotes more stable and discriminative representations, enhancing the model's ability to generalize and improving overall performance.
\end{remark}

\begin{table*}[t]
\renewcommand{\arraystretch}{1.0}
  \centering
  \caption{Dataset statistics for node classification and clustering.}
  \label{tab1}
  \begin{tabular}{l l l l l l l l}
    \toprule
    \makebox[0.2\textwidth][l]{Datasets} & \makebox[0.25\textwidth][l]{Category} & \makebox[0.1\textwidth][l]{Node}
                                     & \makebox[0.1\textwidth][l]{Edge} & \makebox[0.1\textwidth][l]{Feature} & \makebox[0.1\textwidth][l]{Class}            \\

    \midrule
    {Cora \cite{yang2016revisiting}} & Citation Network & 2708 & 5429 & 1433 & 7 \\
    {CiteSeer \cite{yang2016revisiting}} & Citation Network & 3327 & 4732 & 3703 & 6\\
    {PubMed \cite{yang2016revisiting}} & Citation Network & 19717 & 44338 & 500 & 3 \\
    {Computers \cite{suresh2021adversarial}} & Purchase & 13752 & 245861 & 767 & 10  \\
    {Photo \cite{suresh2021adversarial}} & Purchase  & 7650 & 119081 & 745 & 8 \\		
    {CS \cite{suresh2021adversarial}} & Citation Network  & 18333 & 81894 & 6805 & 15 \\	
    {Physics \cite{suresh2021adversarial}} & Citation Network  & 34493 & 247962 & 8415 & 5 \\
    {WikiCS \cite{mernyei2020wiki}} & Citation Network  & 11701 & 216123 & 300 & 10 \\
    \bottomrule
  \end{tabular}
\end{table*}

\section{Experimental Evaluation}
\label{Experimental Evaluation}
\subsection{Experimental Setting}
\textbf{Datasets.} We evaluate the performance of the proposed AS-GCL method by comparing it to that of existing state-of-the-art methods across eight real datasets: Cora, CiteSeer, PubMed, Amazon-Photo, Amazon-Computers, Coauthor-CS, Coauthor-Physics, and WikiCS. The characteristics of these datasets are summarized in TABLE \ref{tab1}.
Cora, CiteSeer, and PubMed are well-known citation networks where nodes represent documents and edges denote citation links between documents.
Amazon-Photo (Photo) and Amazon-Computers (Computers) are subsets of the Amazon co-purchasing graph. In these datasets, nodes represent products, and edges indicate frequently co-purchased items.
Coauthor-CS (CS) and Coauthor-Physics (Physics) are co-authorship graphs derived from academic papers in computer science and physics, respectively, wherein the nodes represent authors, and the edges signify collaborative relationships.
WikiCS is a dataset derived from Wikipedia articles, with the nodes representing articles and the edges reflecting hyperlinks between articles.
These datasets encompass a diverse range of graph-based structures, providing a broad testing ground for evaluating the effectiveness of our method.

\textbf{Baselines.} We examine a range of baseline methods for node-level tasks. These include random walk-based methods such as DeepWalk \cite{perozzi2014deepwalk} and Node2Vec \cite{grover2016node2vec}, as well as supervised learning techniques like MLP, GCN \cite{kipf2016semi}, and GAT \cite{velivckovic2017graph}. We also investigate several self-supervised learning baseline methods categorized into three groups based on their methodology: random consistency perturbations (e.g., DGI \cite{velivckovic2018deep}, GMI \cite{peng2020graph}, MVGRL \cite{hassani2020contrastive}, GRACE \cite{zhu2021deep}, and GraphCL \cite{you2021graph}), generative methods (e.g., VGAE \cite{kipf2016variational} and AdaGCL \cite{jiang2023adaptive}), and learnable techniques such as GCA-SSG \cite{zhang2021canonical}, GRADE \cite{wang2022uncovering}, NCLA \cite{shen2023neighbor}, GCIL \cite{mo2024graph}, and LSGCL \cite{10286310}.
For node clustering tasks, we performed experiments using a variety of clustering and graph self-supervised learning methods. The clustering methods include $K$-means, spectral clustering, GAE \cite{kipf2016variational}, VGAE \cite{kipf2016variational}, DGI \cite{velivckovic2018deep}, DNGR \cite{cao2016deep}, TADW \cite{yang2015network}, and GC-VAE \cite{guo2022graph}, and the self-supervised learning methods are NCLA \cite{shen2023neighbor} and GCIL \cite{mo2024graph}.

\textbf{Evaluation protocol.} To assess the performance of AS-GCL, we utilize a two-layer GCN model where each layer has a hidden dimension of 256. A linear classifier is employed at the postprocessing stage for evaluation. For spectral augmentation, we set the number of training rounds to five and fix the edge perturbation rate \(\epsilon\) at 0.2. During contrastive learning, we run 1000 training epochs with a batch size of 128. The diffusion layers are set as \(i = 2\) and \(k = 2\).
The hyperparameters \(\alpha\) and \(\beta\) are tuned for each dataset, with \(\alpha\) ranging from 4-6 and \(\beta\) ranging from 8-10. For the downstream classification task, we adopt the cross-entropy loss function. The dataset is split as follows: 10\% for training, 10\% for validation, and 80\% for testing.
We use the Adam optimizer with a learning rate of 0.001 and a weight decay of \(5 \times 10^{-5}\) for optimization. To ensure unbiased results, each dataset is run five times with different random seeds, and we report the average accuracy and standard deviation.
All the experiments are performed via PyTorch on a server equipped with eight NVIDIA GeForce 3090 GPUs, each with 24 GB of memory.

\begin{table*}[t]
\renewcommand{\arraystretch}{1.0}
  \centering
  \caption{Accuracy (\%) on the eight benchmark datasets for the node classification task. The best result is represented by boldface font, and the second best result is underlined.}
  \label{tab2}
  \resizebox{\textwidth}{!}{%
  \begin{tabular}{cccccccccc}
    \toprule
    \multirow{2}{*}{\textbf{Method}} & \multicolumn{9}{c}{\textbf{Datasets}}
    \\
    \cline{2-10}
    &Cora&CiteSeer&PubMed&Computers&Photo&CS&Physics&WikiCS&Avg.Acc.\\
    \midrule
   {Deep Walk} &69.1{$\pm$0.3} &43.5{$\pm$0.2} &65.1{$\pm$0.4} &85.7{$\pm$0.1} &89.5{$\pm$0.1} &84.2{$\pm$0.4} &91.7{$\pm$0.2} &74.4{$\pm$0.2}  &75.4\\
   {Node2cec} &71.0{$\pm$0.6} &47.3{$\pm$0.5} &66.3{$\pm$0.8}  &84.6{$\pm$0.2}  &89.7{$\pm$0.2}  &85.2{$\pm$0.1}   &91.2{$\pm$0.1}  &71.8{$\pm$0.1} &75.9\\
    \midrule
   {MLP} &49.9{$\pm$0.4} &49.3{$\pm$0.3} &69.1{$\pm$0.3} &73.9{$\pm$0.2} &79.5{$\pm$0.3} &90.3{$\pm$0.2} &93.5{$\pm$0.4} &72.0{$\pm$0.4} &72.2 \\
   {GCN} &81.6{$\pm$0.2} &70.3{$\pm$0.4} &79.3{$\pm$0.2} &84.5{$\pm$0.3} &91.6{$\pm$0.3} &\underline{93.1{$\pm$0.3}} &93.7{$\pm$0.2} &{73.0{$\pm$0.1}} &83.4\\
   {GAT} &83.1{$\pm$0.3} &72.4{$\pm$0.3} &79.5{$\pm$0.1} &85.8{$\pm$0.1} &91.7{$\pm$0.2} &89.5{$\pm$0.3} &93.5{$\pm$0.3} &{72.6{$\pm$0.3}} &83.5 \\
    \midrule
   {DGI} &82.3{$\pm$0.3} &71.3{$\pm$0.4} &79.4{$\pm$0.3} &84.9{$\pm$0.3} &91.6{$\pm$0.1} &92.1{$\pm$0.4} &93.7{$\pm$0.2} &75.3{$\pm$0.1}  &83.8\\
   {GMI} &83.3{$\pm$0.2} &72.6{$\pm$0.2} &79.8{$\pm$0.4} &82.2{$\pm$0.1} &90.7{$\pm$0.2} &92.6{$\pm$0.2} &{94.3{$\pm$0.4}} &74.9{$\pm$0.2} &83.8\\		
   {VGAE} &75.9{$\pm$0.5} &66.8{$\pm$0.2} &75.8{$\pm$0.4} &85.8{$\pm$0.3} &91.5{$\pm$0.2} &91.8{$\pm$0.4} &94.1{$\pm$0.2} &75.5{$\pm$0.3} &82.2\\	
   {MVGRL} &83.1{$\pm$0.2} &72.3{$\pm$0.5} &80.3{$\pm$0.5} &87.5{$\pm$0.1} &91.7{$\pm$0.1} &92.1{$\pm$0.3} &\underline{95.1{$\pm$0.2}} &77.5{$\pm$0.1} &84.9\\
   {GRACE} &81.9{$\pm$0.6} &71.4{$\pm$0.4} &80.5{$\pm$0.4} &86.3{$\pm$0.5} &92.1{$\pm$0.3} &84.2{$\pm$0.4} &92.9{$\pm$0.2} &78.1{$\pm$0.1} &83.4\\

   {GraphCL} &82.1{$\pm$0.3} &72.8{$\pm$0.2} &81.7{$\pm$0.3} &85.6{$\pm$0.2} &92.5{$\pm$0.1} &92.6{$\pm$0.4} &93.7{$\pm$0.2} &76.8{$\pm$0.3} &84.7\\
   {GCA-SSG} &83.9{$\pm$0.4} &\underline{73.1{$\pm$0.3}} &81.3{$\pm$0.4} &\underline{88.4{$\pm$0.3}} &89.5{$\pm$0.1} &92.4{$\pm$0.1} &93.4{$\pm$0.2} &78.2{$\pm$0.3} &85.0\\
   {GRADE} &{84.0{$\pm$0.3}} &72.4{$\pm$0.4} &{82.7{$\pm$0.3}} &84.7{$\pm$0.1} &\underline{92.6{$\pm$0.1}} &92.7{$\pm$0.4} &93.7{$\pm$0.2} &78.1{$\pm$0.2} &85.1\\
  
{NCLA} &{82.2{$\pm$1.6}} &71.7{$\pm$0.9} &{82.0{$\pm$1.4}} &83.7{$\pm$1.1} &{90.2{$\pm$1.3}} &91.5{$\pm$0.7} &92.2{$\pm$0.9} &{78.6{$\pm$0.8}} &84.0\\
{AdaGCL} &{83.8{$\pm$1.2}} &70.5{$\pm$0.6} &{82.6{$\pm$0.5}} &84.3{$\pm$1.0} &{91.2{$\pm$0.8}} &90.4{$\pm$0.6} &92.6{$\pm$0.7} &{77.4{$\pm$0.5}} &84.1\\
{GCIL} &{84.2{$\pm$0.5}} &69.1{$\pm$0.4} &{81.6{$\pm$0.7}} &83.4{$\pm$0.6} &{90.5{$\pm$0.4}} &91.8{$\pm$0.9} &92.8{$\pm$0.4} &{76.7{$\pm$0.6}} &83.8\\
 {LSGCL} &\underline{84.9{$\pm$0.3}} &72.0{$\pm$0.4} &\underline{84.5{$\pm$0.3}} &85.6{$\pm$0.1} &{91.5{$\pm$0.2}} &91.8{$\pm$0.2} &92.9{$\pm$0.2} &\underline{79.1{$\pm$0.3}} &\underline{85.2}\\
    \midrule
   {Ours} &\textbf{85.2{$\pm$0.3}} &\textbf{73.9{$\pm$0.1}} &\textbf{84.9{$\pm$0.3}} &\textbf{89.9{$\pm$0.3}} &\textbf{93.5{$\pm$0.2}} &\textbf{93.5{$\pm$0.3}} &\textbf{95.2{$\pm$0.2}} &\textbf{79.5{$\pm$0.1}} &\textbf{86.9}\\
    \bottomrule
  \end{tabular}
  }
\end{table*}

\begin{figure}[t]
\centering
\includegraphics[width=0.45\columnwidth]{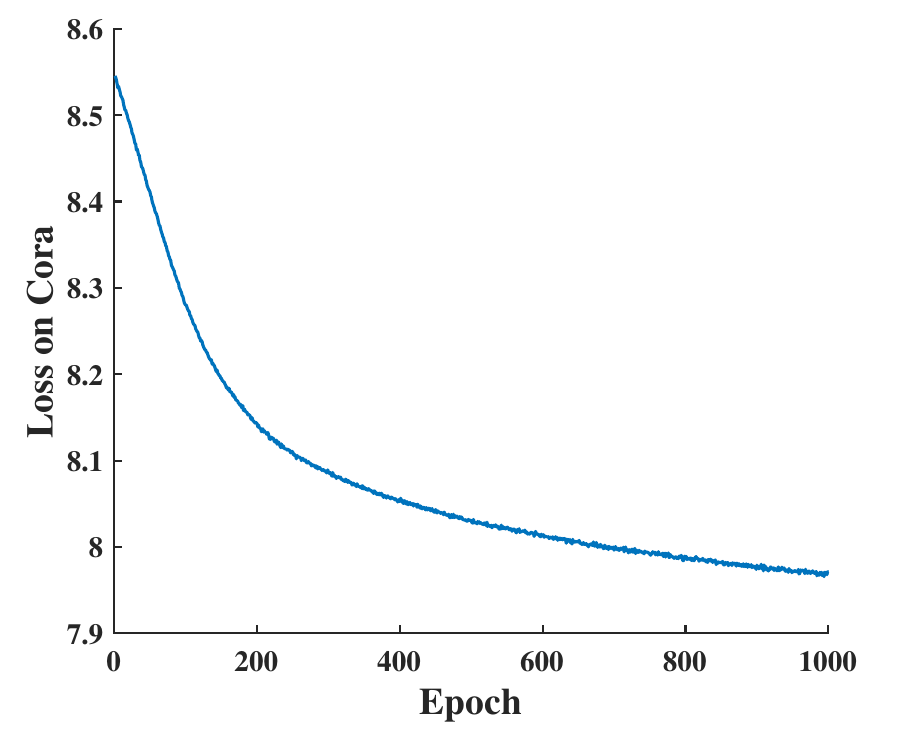}
\hfill
\includegraphics[width=0.45\columnwidth]{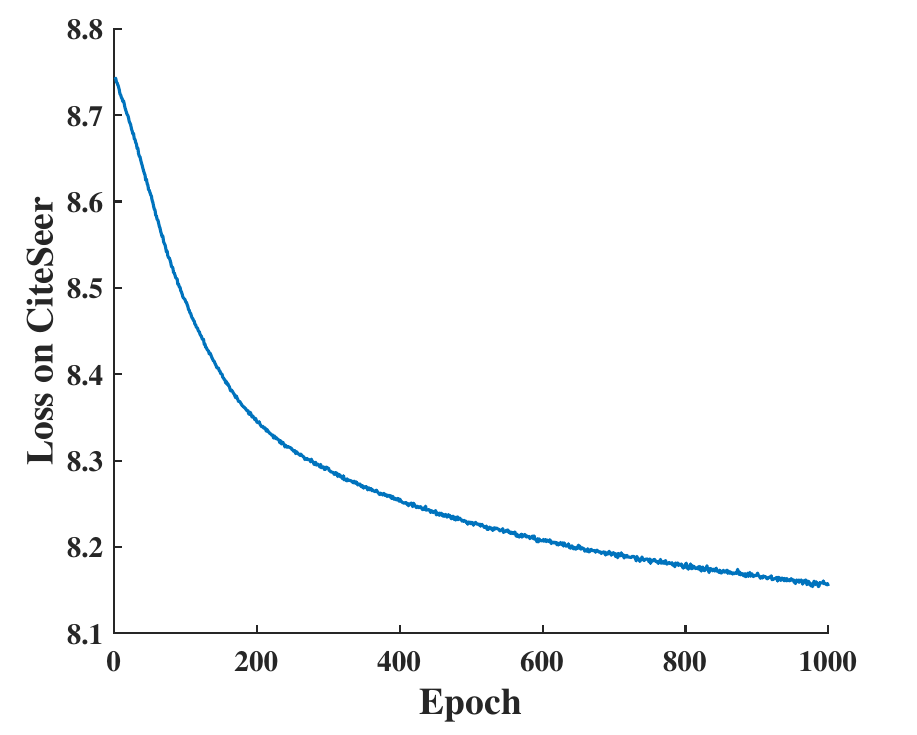}
\caption{Training loss for different numbers of rounds.}
\label{fig11}
\end{figure}

\subsection{Experimental Results}
\textbf{Node Classification.}
TABLE \ref{tab2} presents the node classification accuracy results, where the best-performing method in each column is highlighted in boldface font, and the second-best method is underlined. Fig. \ref{fig11} depicts the training loss variation over the number of training rounds on the Cora and CiteSeer datasets. Notably, AS-GCL achieves state-of-the-art (SOTA) performance across seventeen graph benchmarks. Compared with other self-supervised approaches applied to the eight datasets, AS-GCL outperforms the strongest baseline, LSGCL, by an average of 1.7\%, and surpasses the weakest baseline, VGAE, by 4.7\%. These results demonstrate the effectiveness of AS-GCL  on node classification tasks, further highlighting its ability to handle diverse datasets and graph structures with  exceptional accuracy.

\begin{table*}[t]
\renewcommand{\arraystretch}{1.0}
  \centering
  \caption{ACC, NMI, F-score and ARI on three benchmark datasets for the node clustering task. The best result is in boldface font, and the second best result is underlined.}
  \label{tab3}
  \resizebox{\textwidth}{!}{%
  \begin{tabular}{c|c|cccccccccc|c}
    \toprule
    \multirow{2}{*}{\textbf{Datasets}} & \multicolumn{12}{c}{\textbf{Methods}} \\
    \cline{2-13}
    & \small{Metirc} & \small{K-Means} & \small{Spectral} & \small{GAE} &\small{VGAE} & \small{DGI} & \small{DNGR} & \small{TADW} & \small{GC-VAE}  &\small{NCLA} &\small{GCIL} &\small{Ours} \\
    \midrule
    \multirow{4}{*}{{Cora}}
    &\small{ACC} &\small{0.493} &\small{0.396} &\small{0.597} &\small{0.592} &\small{0.590} &\small{0.419} &\small{0.562} &\small{\underline{0.707}} &\small{0.694} &\small{0.688}&\small{\textbf{0.728}}  \\
    &\small{NMI} &\small{0.311} &\small{0.289} &\small{0.392} &\small{0.408} &\small{0.386} &\small{0.317} &\small{0.441} &\small{0.537} &\small{0.543} &\small{\underline{0.552}}&\small{\textbf{0.584}}  \\
    &\small{F-score} &\small{0.376} &\small{0.332} &\small{0.415} &\small{0.456} &\small{0.432} &\small{0.389} &\small{0.418} &\small{\textbf{0.695}}  &\small{0.485}  &\small{0.521} &\small{\underline{0.683}} \\
    &\small{ARI} &\small{0.230} &\small{0.176} &\small{0.294} &\small{0.342} &\small{0.336} &\small{0.142} &\small{0.332} &\small{\underline{0.482}} &\small{0.425}  &\small{0.414}&\small{\textbf{0.491}}  \\
    \midrule
    \multirow{4}{*}{{CiteSeer}}
     &\small{ACC} &\small{0.541} &\small{0.318} &\small{0.413} &\small{0.603} &\small{0.577} &\small{0.326} &\small{0.455} &\small{\underline{0.663}} &\small{0.485}  &\small{0.623}&\small{\textbf{0.671}}  \\
    &\small{NMI} &\small{0.315} &\small{0.087} &\small{0.174} &\small{0.343} &\small{0.319} &\small{0.182} &\small{0.291} &\small{0.407} &\small{0.394} &\small{\underline{0.409}}  &\small{\textbf{0.417}}  \\
    &\small{F-score} &\small{0.413} &\small{0.256} &\small{0.297} &\small{0.462} &\small{0.452} &\small{0.403} &\small{0.415} &\small{\underline{0.632}} &\small{0.505} &\small{0.548}&\small{\textbf{0.651}} \\
    &\small{ARI} &\small{0.279} &\small{0.084} &\small{0.143} &\small{0.343} &\small{0.289} &\small{0.047} &\small{0.354} &\small{\underline{0.415}} &\small{0.402} &\small{0.394} &\small{\textbf{0.417}}  \\
    \midrule
    \multirow{4}{*}{{PubMed}}
     &\small{ACC} &\small{0.562} &\small{0.498} &\small{0.608} &\small{0.619} &\small{0.499} &\small{0.468} &\small{0.355} &\small{\underline{0.682}} &\small{0.532} &\small{0.487} &\small{\textbf{0.687}}   \\
    &\small{NMI} &\small{0.262} &\small{0.142} &\small{0.235} &\small{0.213} &\small{0.151} &\small{0.155} &\small{0.103} &\small{\underline{0.294}} &\small{0.208}  &\small{0.253}  &\small{\textbf{0.296}}  \\
    &\small{F-score} &\small{0.559} &\small{0.473} &\small{0.497} &\small{0.478} &\small{0.432} &\small{0.452} &\small{0.426} &\small{\underline{0.663}} &\small{0.452} &\small{0.424} &\small{\textbf{0.669}}  \\
    &\small{ARI} &\small{0.227} &\small{0.098} &\small{0.223} &\small{0.211} &\small{0.145} &\small{0.054} &\small{0.158} &\small{{0.273}}  &\small{{0.252}}  &\small{\underline{0.277}} &\small{\textbf{0.297}}  \\
    \bottomrule
  \end{tabular}
}
\end{table*}

\textbf{Node Clustering.}
Additionally, we evaluate the impact of the proposed method on node clustering by conducting a comprehensive performance analysis on three benchmark datasets: Cora, Citeseer, and Pubmed. The results, detailed in TABLE \ref{tab3}, underscore the exceptional performance of AS-GCL compared with that of state-of-the-art graph clustering methods. The highest results in each column are highlighted in boldface font, and the second-highest results are underlined. AS-GCL consistently delivers strong results across multiple evaluation metrics, including Accuracy (ACC), Normalized Mutual Information (NMI), F-measure (F-score), and Adjusted Rand Index (ARI). Such consistent improvements across different benchmarks and metrics further validate the robustness and effectiveness of AS-GCL on node clustering tasks, highlighting its potential for broader applications in graph-based learning.

\textbf{Analysis.}
The excellent performance of AS-GCL is largely attributable to several key advantages of the proposed method. First, our spectral-based augmentation strategy plays a critical role in the enhanced performance of AS-GCL. By leveraging spectral information, AS-GCL captures both global and local structural patterns in graphs, ensuring more informative and diverse contrastive views. This enables the model to preserve essential graph properties while effectively reducing noise. Second, the asymmetric encoder design enhances learning by employing parameter-sharing encoders with different diffusion operators. This approach not only increases the diversity of the learned graph representations but also mitigates overfitting. As a result, AS-GCL is 
highly 
equipped to generalize across different datasets. Third, the upper-bound loss function incorporated in AS-GCL ensures balanced intraclass and interclass distance. By addressing the issue of excessive intraclass and insufficient interclass distance, our method enhances the discriminative power of learned representations, leading to improved classification performance.

\subsection{Empirical Running Time}
The time complexity for precomputing each round of spectral augmentation is $\mathcal{O}(mnT)$, where $m$ represents the number of edges, $n$ is the number of nodes, and $T$ denotes the number of augmentation rounds. Typically, $T$ is a small constant, often set to less than 5, making the overall precomputation cost manageable. TABLE \ref{tab_time} shows the runtime (in seconds) of the proposed method along with those of typical graph contrastive learning methods. In these experiments, we performed 1000 training iterations on datasets of varying size, conducting five rounds of iterative training for spectral augmentation for our method. The precomputational overhead associated with our spectral augmentation method is reasonable and acceptable, especially in the context of overall runtime; this is because our method requires only a single data enhancement step before initiating formal training. Additionally, our approach demonstrates outstanding classification and clustering performance with time efficiency comparable to that of generative methods (see TABLE \ref{tab2} and TABLE \ref{tab3}). For large-scale graphs, where time and memory complexity are critical, techniques such as Chebyshev polynomial approximation of the graph Laplacian and subgraph sampling can further reduce the computational burden. These strategies enable more efficient batch training and enhance scalability. We emphasize that our work represents a preliminary step toward effective graph augmentation, with further research needed to scale the method to even larger graphs, a key focus for future investigation.

\begin{table}[t]
\renewcommand{\arraystretch}{1.0}
  \centering
  \caption{Comparisons of empirical runtime (in seconds) on four representative node classification datasets with varying number of nodes. In AS-GCL, `Overall runtime' denotes the sum of `Data augmentation time' and `Contrastive training time' for comparison with other baseline methods.}
  \label{tab_time}
  \resizebox{0.48\textwidth}{!}{%
    \begin{tabular}{llllll}
\toprule
\multicolumn{2}{l}{Dataset}                                                                  & Cora  & Photo  & Computers & PubMed \\ \midrule
\multicolumn{2}{l}{Number of nodes}                                                          & 2708  & 7650   & 13752     & 19717  \\ \midrule
\multicolumn{1}{c}{\multirow{3}{*}{AS-GCL (Ours)}} & Data augmentation time (epoch = 5)   & 96.4  & 472.4  & 870.3     & 1235.7 \\
\multicolumn{1}{c}{}                              & Contrastive training time (epoch = 1000) & 257.8 & 838.2  & 1103.0    & 1389.2 \\
\multicolumn{1}{c}{}                              & Overall runtime           & 354.2 & 1310.6 & 1973.3    & 2624.9 \\ \midrule
\multicolumn{2}{l}{GraphCL}                                                                  & 289.4 & 1028.5 & 1531.7    & 1864.2 \\
\multicolumn{2}{l}{VGAE}                                                                     & 314.6 & 1205.3 & 1784.4    & 2287.3 \\
\multicolumn{2}{l}{AdaGCL}                                                                   & 364.3 & 1425.1 & 1967.4    & 2753.8 \\
\multicolumn{2}{l}{NCLA}                                                                     & 342.0 & 1152.3 & 1684.7    & 2432.5 \\
\multicolumn{2}{l}{GCIL}                                                                        & 372.5 & 1422.7 & 2008.6 & 2542.2        \\ \bottomrule
\end{tabular}
  }
\end{table}

\subsection{Spectral Variation Verification}
\begin{figure}[t]
\centering
\includegraphics[width=0.45\columnwidth]{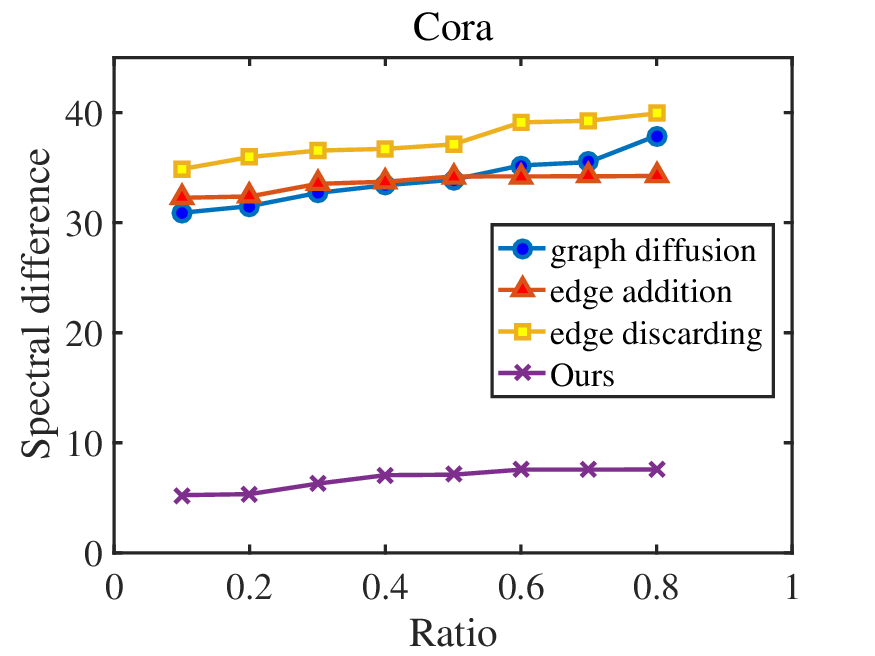}
\hfill
\includegraphics[width=0.45\columnwidth]{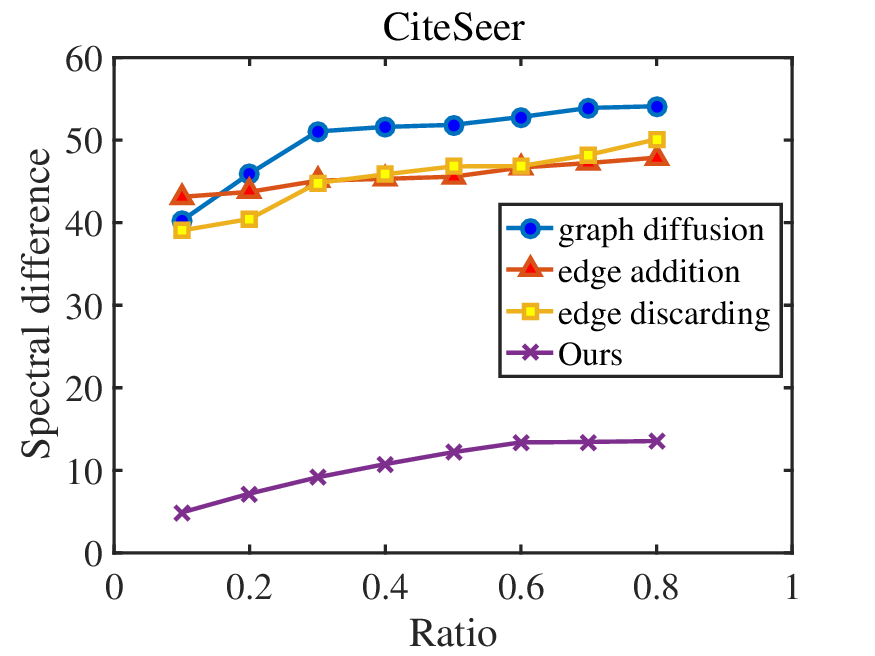}
\caption{Frobenius distance between the symmetric normalized Laplace matrices of the decomposition of the original and augmented graphs.}
\label{fig2}
\end{figure}
We can quantify the changes in Laplace operators resulting from graph augmentation. To assess the impact of graph augmentation on the components of the adjacency matrix, we denote the symmetric normalized Laplace operator of the original graph as \(\hat{\boldsymbol{L}}\) and the symmetric normalized Laplace operator of the augmented graph as \(\widetilde{\boldsymbol{L}}\). We use the Frobenius norm \(\|\hat{\boldsymbol{L}} - \widetilde{\boldsymbol{L}}\|_F\) as a metric to measure the distance between these operators. In addition, we examine the effects of three commonly employed augmentations on the adjacency matrix: edge addition, edge removal, and graph diffusion. The Frobenius distances between the symmetrically normalized Laplacian matrices of the augmented and original graphs are summarized in Fig. \ref{fig2}. Notably, significant alterations in the graph Laplace operator result from all three random topological augmentations; t
his highlights the limitations of using consistent random edge perturbations to preserve the structural properties of graphs. 
The proposed method addresses this limitation by using minimal spectral variations as a guide. As a result, the proposed method effectively captures the invariant spectral components associated with sensitive edges while reducing its reliance on unstable components, proposing a more efficient representation of the graph.

\subsection{Ablation Studies}
We conduct an ablation study to assess the individual contributions of various components within the unified learning framework of AS-GCL. The framework includes several key steps, i.e., graph data augmentation, view encoding, and contrastive loss computation. In particular, we focus on three crucial components: spectral augmentation for optimizing stochastic topology augmentation, the asymmetric encoder for enhancing view diversity and reducing noise, and the use of upper-bound and lower-bound loss to minimize intraclass distance. By systematically removing each component, we can evaluate the impact of each component on overall performance. In TABLE \ref{tab4}, a decline in performance  results when any of the components are omitted, highlighting the significance of each component in improving the model's effectiveness. Both the optimized data augmentation approach and the mechanisms designed to enhance model robustness play a substantial role in AS-GCL's performance, underscoring the importance of integrating all of the components to successfully address graph representation learning tasks.

\begin{table*}[t]
\renewcommand{\arraystretch}{1.0}
  \centering
  \caption{Results of the ablation study of the proposed AS-GCL method. The best results are shown in boldface font.}
  \label{tab4}
  \resizebox{0.9\textwidth}{!}{%
  \begin{tabular}{ccccccccc}
    \toprule
    \multirow{2}{*}{\textbf{Variant}} & \multicolumn{8}{c}{\textbf{Datasets}}
    \\
    \cline{2-9}
    &Cora&CiteSeer&PubMed&Computers&Photo&CS&Physics&WikiCS\\
     \midrule
   {AS-GCL}  &\textbf{85.2} &\textbf{73.9} &\textbf{84.9} &\textbf{89.2} &\textbf{93.5} &\textbf{93.5} &\textbf{95.2} &\textbf{79.5}  \\
 \midrule
   {w/o spectral augmentation } &80.1 &61.3 &81.6 &84.5 &87.7 &86.4 &92.0 &72.2\\
   {w/o asymmetric encoders} &83.1	&68.5 &82.6	&84.9 &90.9	&90.1  &91.2  &73.3 \\
   {w/o upper-bound loss} &82.4	&63.4 &81.4	&84.8	&90.2	&87.9 &89.6	&76.3 \\
   {w/o lower-bound loss}  &84.6 &69.8	&82.6	&83.0	&90.8	&89.8 &90.1	&72.5\\
   {w/o upper-bound and lower-bound losses} &80.8	&62.2	&80.9	&84.5	&90.4	&86.9	&89.6 &75.3 \\	
    \bottomrule
  \end{tabular}
  }
\end{table*}

\begin{figure}[t]
\centering
\includegraphics[width=0.45\columnwidth]{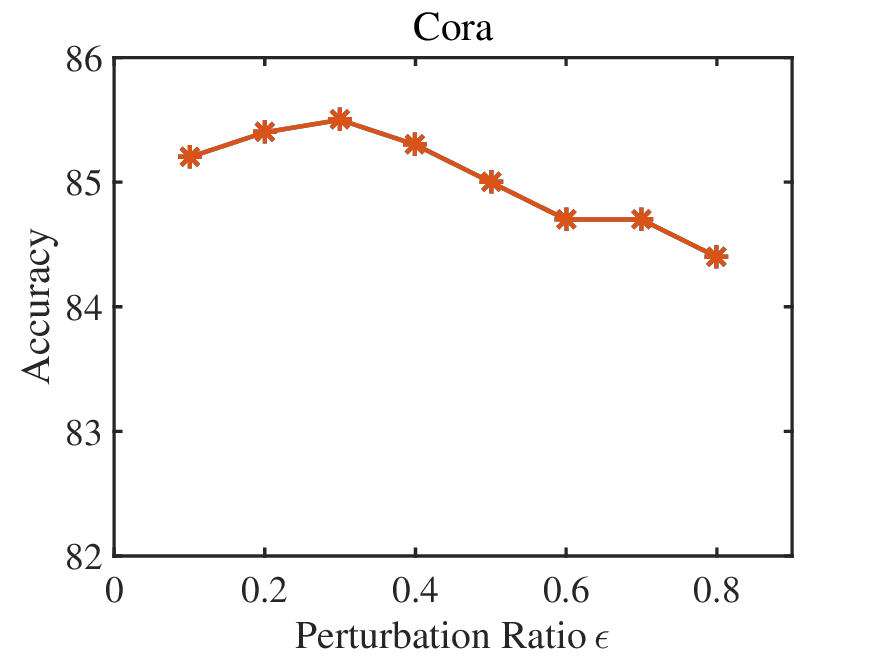}
\hfill
\includegraphics[width=0.45\columnwidth]{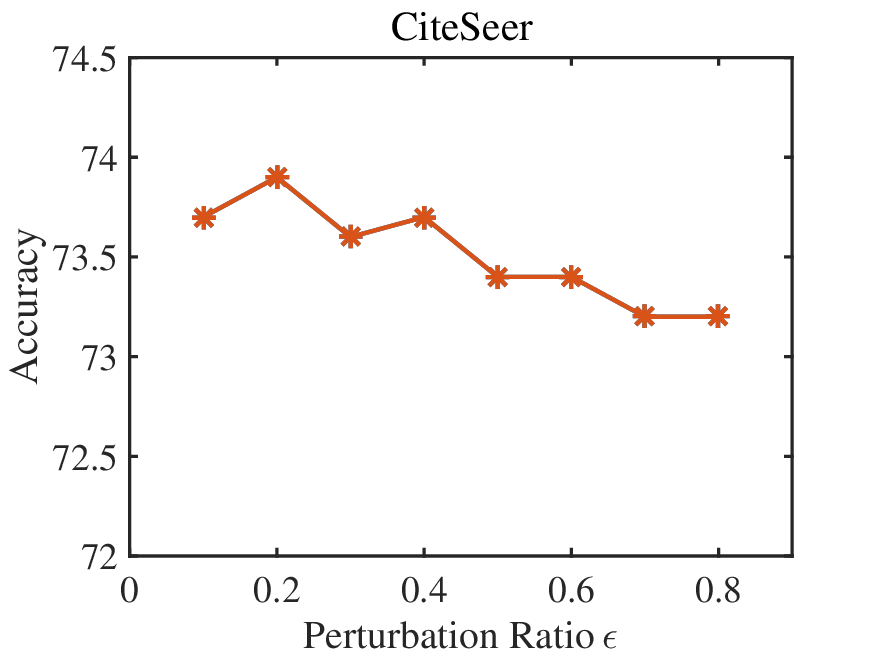}
\caption{Node classification accuracy with different edge perturbation ratios $\epsilon$.}
\label{fig412}
\end{figure}

\begin{figure}[t]
\centering
\includegraphics[width=0.45\columnwidth]{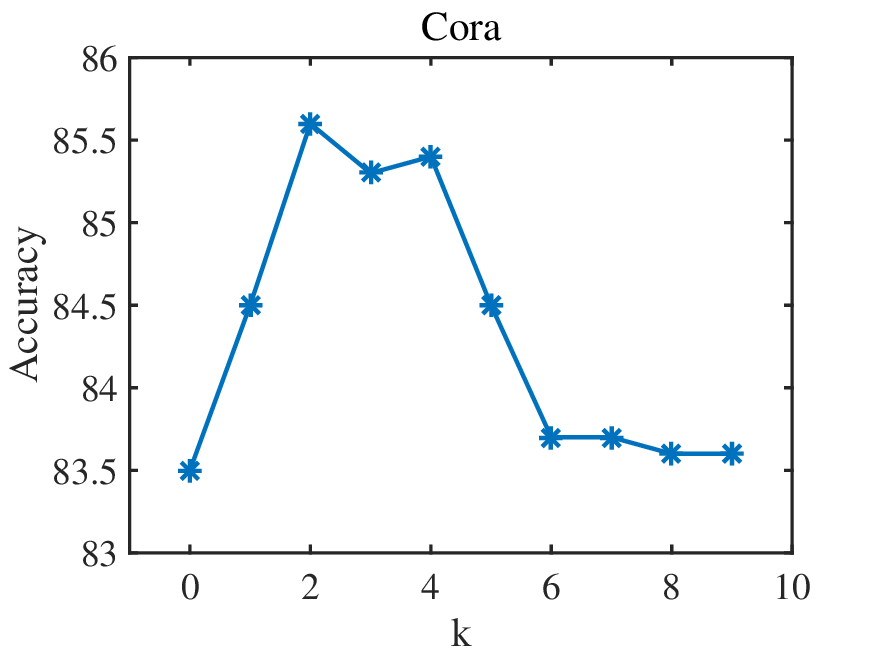}
\hfill
\includegraphics[width=0.45\columnwidth]{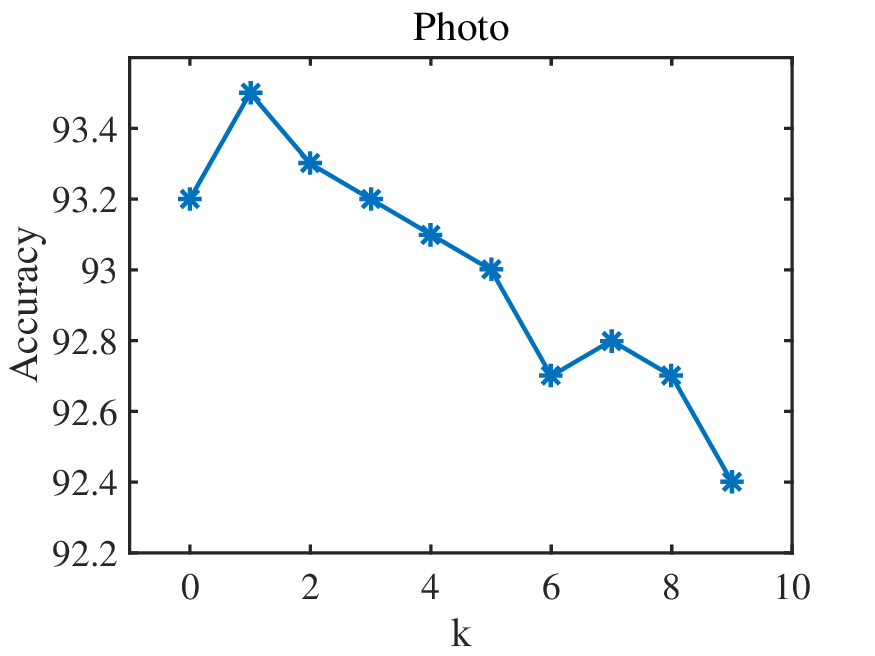}
\caption{Node classification accuracy with different diffusion layers $k$ in asymmetric encoders.}
\label{fig5}
\end{figure}
\begin{figure}[t]
\centering
\includegraphics[width=0.24\textwidth]{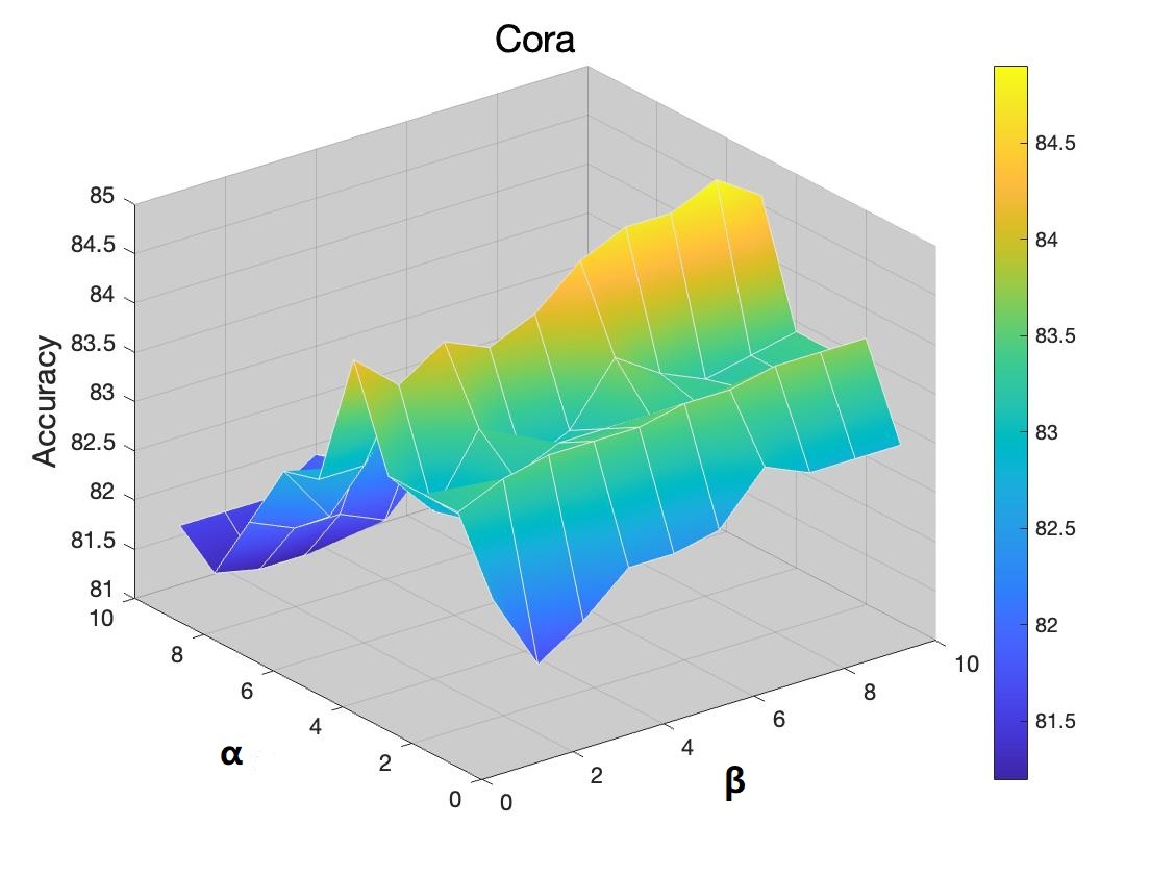}
\includegraphics[width=0.24\textwidth]{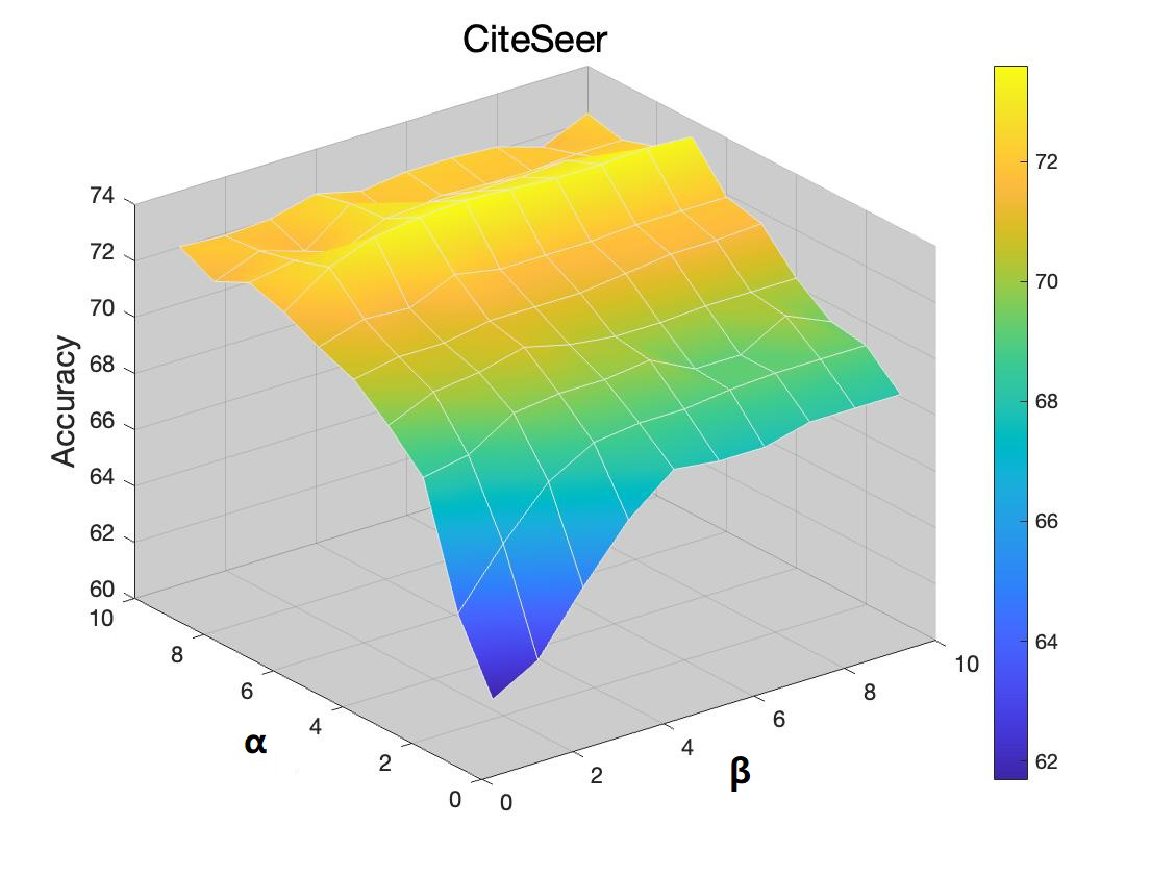}
\includegraphics[width=0.24\textwidth]{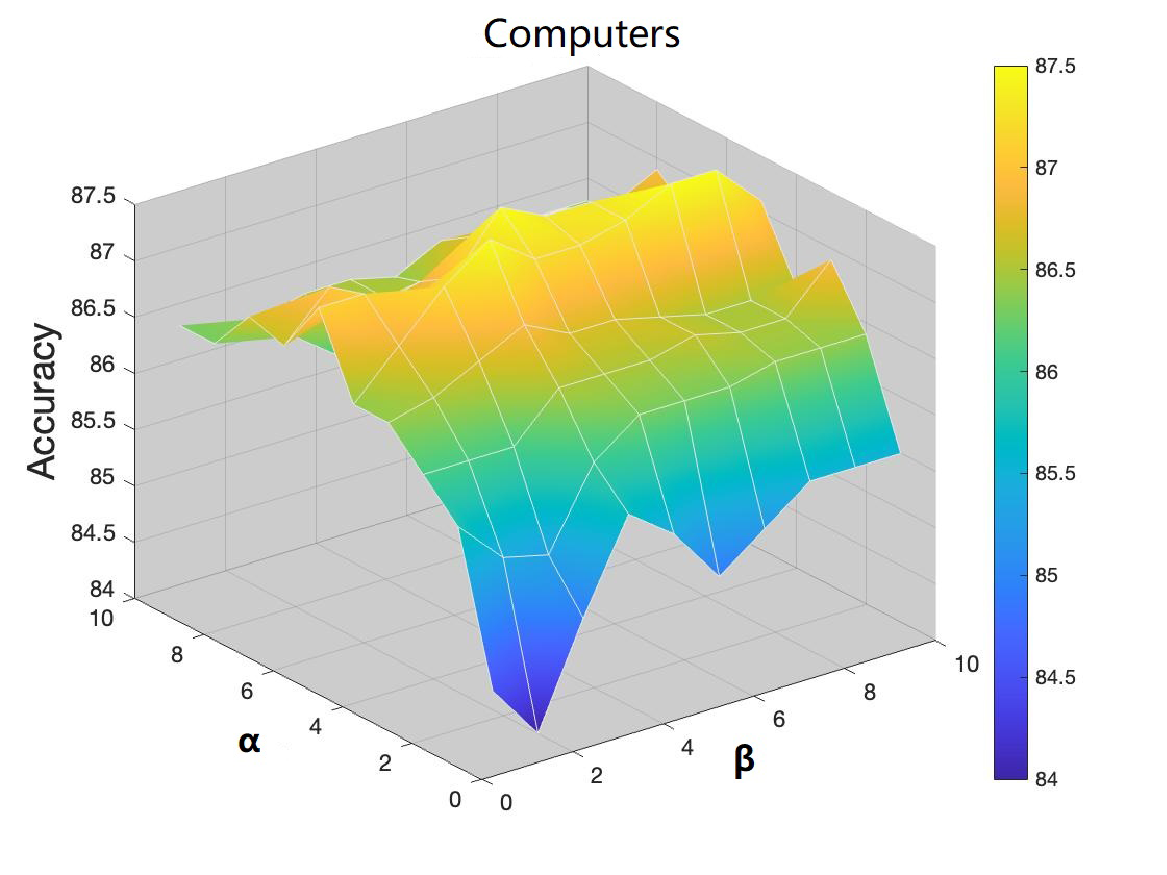}
\includegraphics[width=0.24\textwidth]{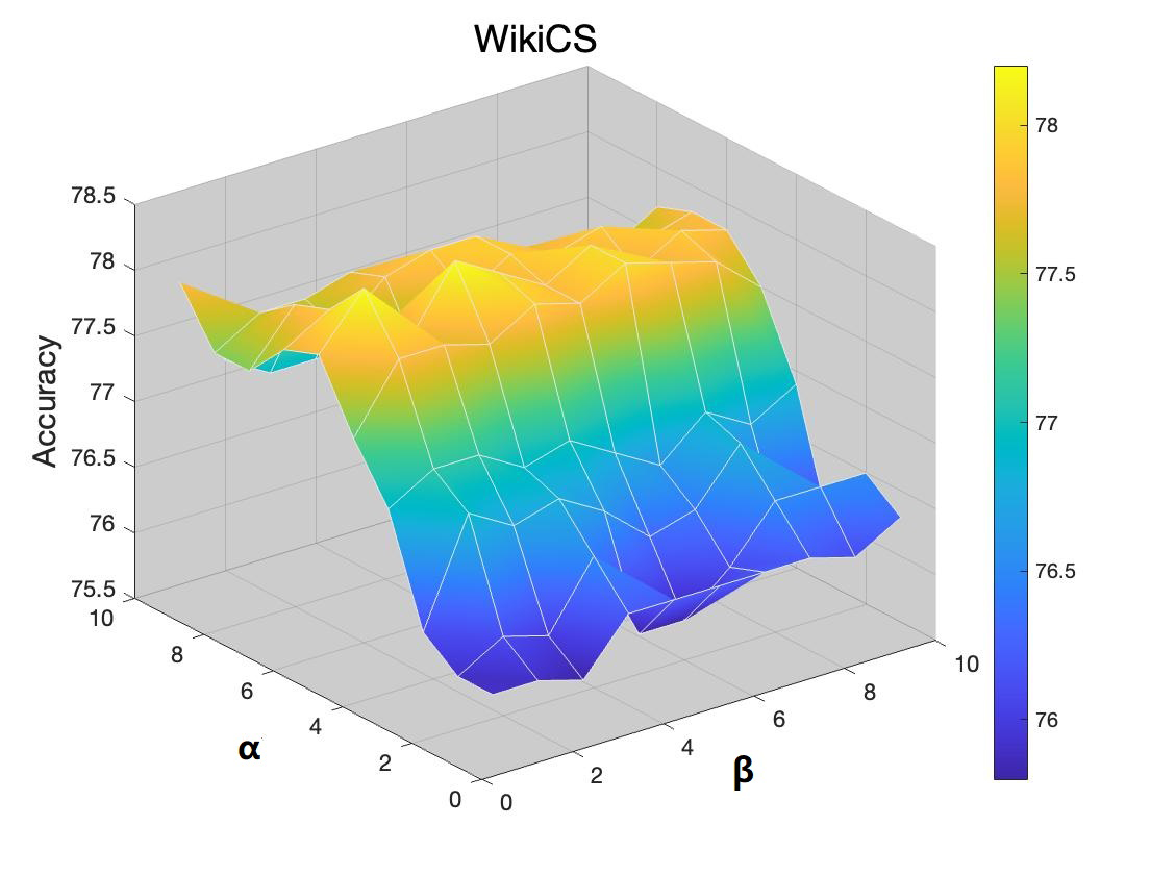}
\caption{Node classification accuracy of the proposed method under different parameter settings ($\alpha$ and $\beta$).}
\label{fig6}
\end{figure}
\subsection{Parametric Experiments}
Next, we evaluate the impact of four key hyperparameters on model performance: perturbation ratio \(\epsilon\), number of diffusion layers in the asymmetric encoder, and thresholds for upper and lower bound loss, denoted as \(\alpha\) and \(\beta\), respectively. As shown in Fig. \ref{fig412}, the model is not highly sensitive to changes in the perturbation ratio. This stability is attributed to the proposed spectral augmentation method, which effectively reduces the influence of noise introduced by perturbation, ensuring that performance remains stable even with increased perturbation levels.

Fig. \ref{fig5} provides insights into the effect of the number of diffusion layers on model performance. With a relatively small number of layers, the asymmetric encoder efficiently filters out high-frequency noise and enhances the retention of important structural information. However, as the number of diffusion layers increases, the information undergoes multiple rounds of propagation, leading to signal smoothing and the loss of key details and local features during aggregation and transmission. This over-smoothing effect ultimately reduces performance, suggesting a need to balance effective information propagation with the preservation of local structural details.

In addition, we investigate the effect of varying the values of \(\alpha\) and \(\beta\) on node classification performance, as illustrated in Fig. \ref{fig6}. Extreme values---either too small or too large---adversely impact the performance of the model. When \(\alpha\) is too small or \(\beta\) is too large, the model's ability to control the distance within and between classes weakens, leading to reduced intraclass compactness and diminished clustering effectiveness. When \(\alpha\) is too large or \(\beta\) is too small, the intraclass distance increases while the interclass distance decreases, which hinders the model's ability to differentiate between distinct classes. Therefore, selecting appropriate values for \(\alpha\) and \(\beta\) is essential to achieving a balanced trade-off between intraclass compactness and interclass separation, which in turn ensures effective classification. The hyperparameter evaluation presented here highlights the flexibility of the AS-GCL method under different configurations.

\begin{figure}[!t]
\centering
\includegraphics[width=0.45\columnwidth]{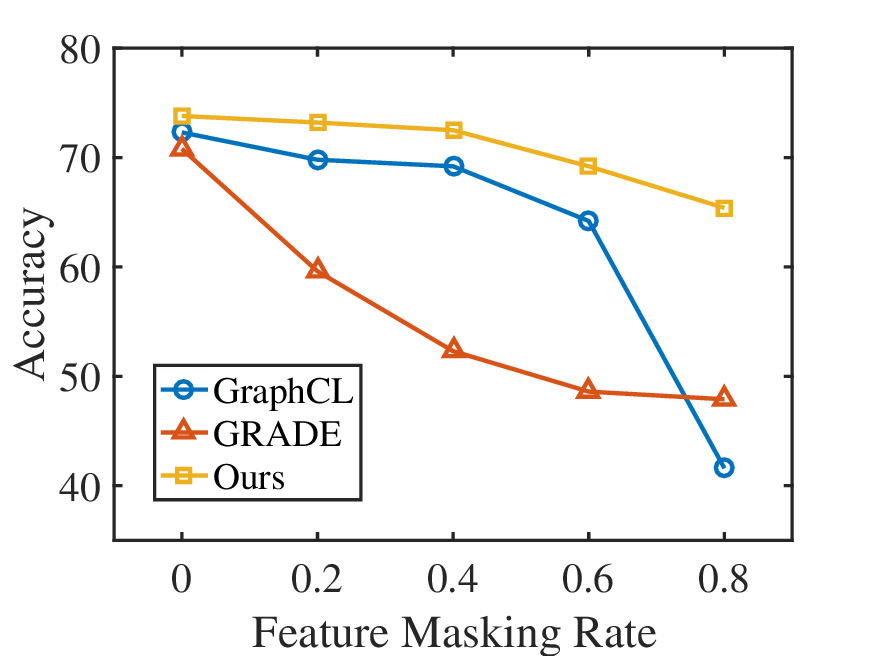}
\hfill
\includegraphics[width=0.45\columnwidth]{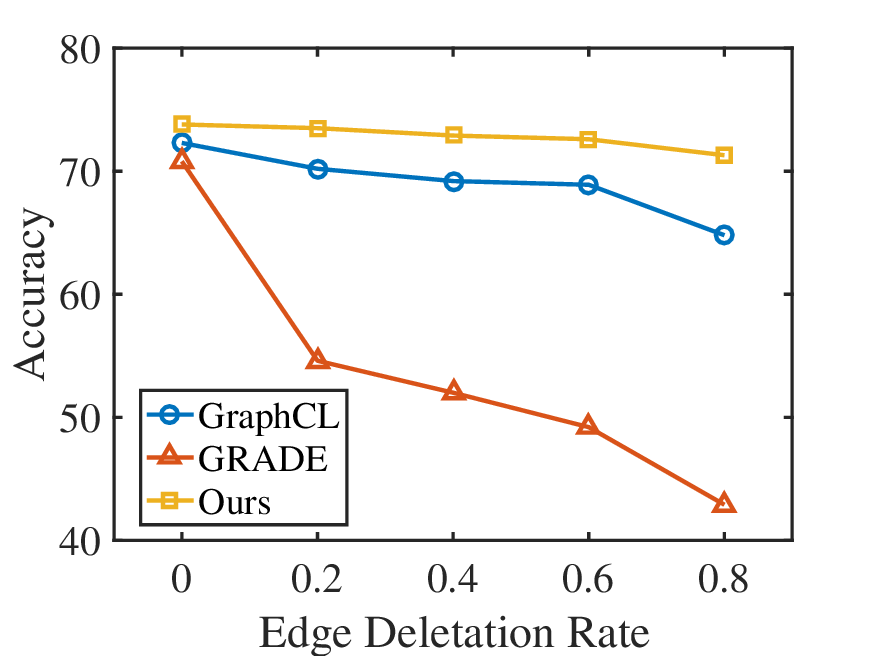}
\caption{Test node classification accuracy when graph structures are perturbed by feature masking or edge deletion.}
\label{fig7}
\end{figure}
\subsection{Robustness Analysis}
To evaluate the robustness of the proposed AS-GCL method against adversarial graph perturbations, we test its performance on the Citeseer dataset under various attack intensities. In this evaluation, we simulate attacks by randomly altering the graph structure through feature masking and edge removal, with the ratio of modified edges and masked features varying from 0-0.8. We compare the performance of AS-GCL against two baseline methods, GraphCL and GRADE. As shown in Fig. \ref{fig7}, AS-GCL consistently delivers excellent or comparable performance in both attack scenarios. Notably, AS-GCL confers significant improvements with higher edge removal rates, demonstrating its robustness in maintaining performance even amidst severe graph perturbations. This adaptability is especially valuable for practical applications where graph structures can be significantly altered or subjected to adversarial changes. The ability of our method to handle substantial structural modifications without considerable performance degradation underscores its effectiveness in robust graph representation learning.

\section{Conclusion}
\label{conclusion}
This paper proposes a novel method called Asymmetric Spectral Augmentation Graph Contrastive Learning (AS-GCL) that introduces a new paradigm for topological augmentation. The proposed method minimizes spectral variations and generates distinct view representations by using encoders with shared parameters but different diffusion operators. This strategy effectively reduces graph structure noise and produces more reliable graph views. Additionally, we introduce an upper-bound loss function to address the significant bias found between positive and anchor embeddings. Extensive experiments on various node-level tasks, including node classification and clustering, demonstrate that AS-GCL outperforms existing methods. Future work will focus on applying AS-GCL to larger datasets and more complex graph neural networks.

\section*{Acknowledgments}
This work is supported in part by the National Natural Science Foundation of China (No. 62106259, No. 62076234), the Beijing Outstanding Young Scientist Program (No. BJJWZYJH012019100020098), and the Beijing Natural Science Foundation (No. 4222029).

\bibliographystyle{IEEEtran}
\bibliography{bare_jrnl_new_sample4}
\vspace{-15mm} 

\begin{IEEEbiography}[{\includegraphics[width=1in,height=1.25in,clip,keepaspectratio]{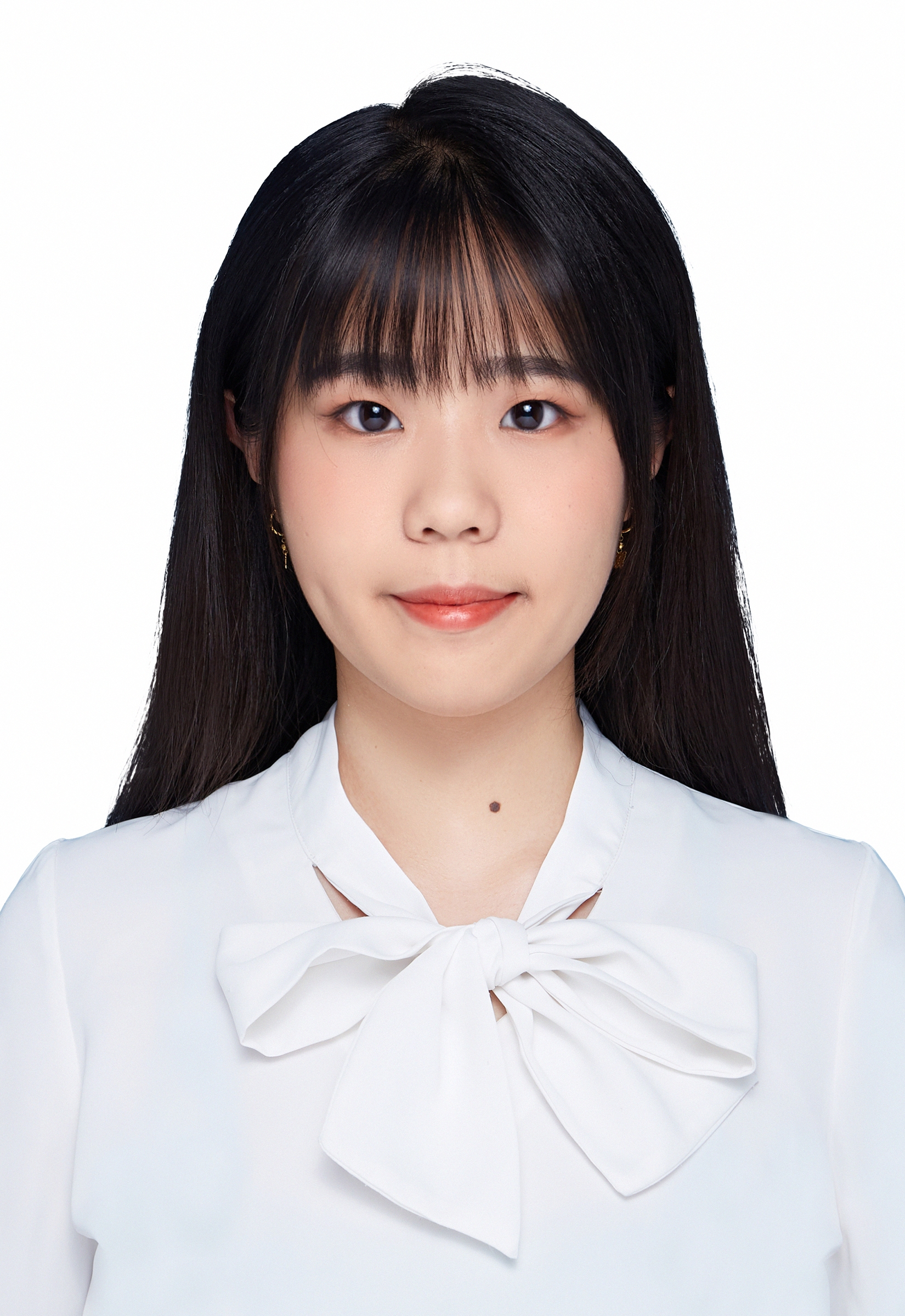}}]{Ruyue Liu}
is a 
Ph.D. 
student at the Institute of Information Engineering, Chinese Academy of Sciences, Beijing, China, and the School of Cyber Security, University of Chinese Academy of Sciences, Beijing, China. Her current research interests include machine learning, data mining, self-supervised learning, federated learning, and graph representation learning.
\end{IEEEbiography}
\vspace{-8mm} 
\begin{IEEEbiography}[{\includegraphics[width=1in,height=1.25in,clip,keepaspectratio]{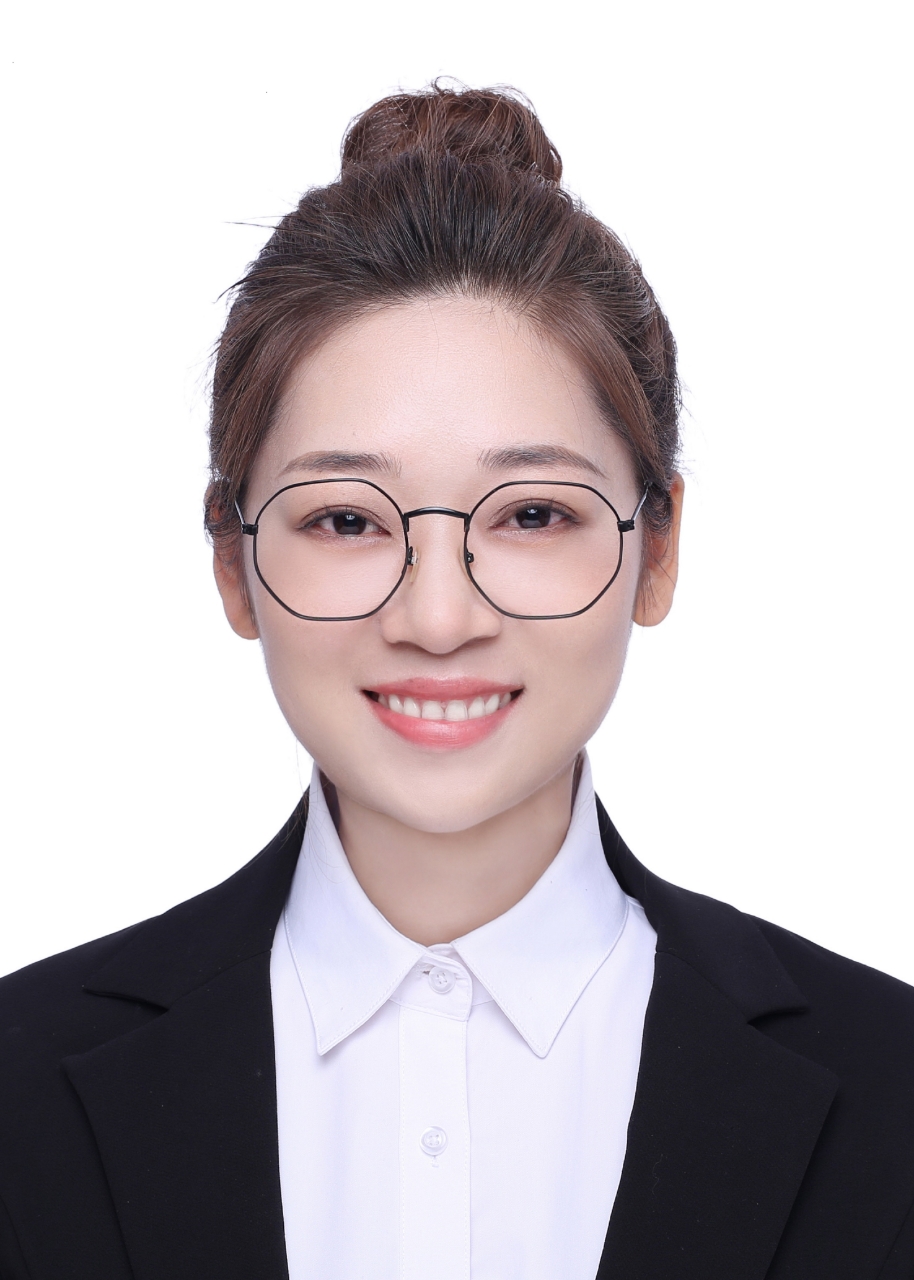}}]{Rong Yin}
received a Ph.D. from the Institute of Information Engineering, Chinese Academy of Sciences, Beijing, China, and the School of Cyber Security, University of Chinese Academy of Sciences, Beijing, China, in 2020. She is currently an Associate Professor with the Institute of Information Engineering, Chinese Academy of Sciences, Beijing, China. Her current research interests include machine learning, data mining, statistical theory, distributed learning, and optimization algorithms.
\end{IEEEbiography}
\vspace{-3 mm}
\begin{IEEEbiography}[{\includegraphics[width=1in,height=1.25in,clip,keepaspectratio]{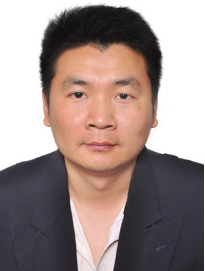}}]{Yong Liu}
was born in 1986. He received a Ph.D. in Computer Science from Tianjin University, Tianjin, China, in 2016. He is currently an Associate Professor with the Beijing Key Laboratory of Big Data Management and Analysis Methods, Gaoling School of Artificial Intelligence, Renmin University of China, Beijing, China. His current research interests include large-scale kernel methods, large-scale model selection, and machine learning.
\end{IEEEbiography}
\vspace{-3 mm}
\begin{IEEEbiography}[{\includegraphics[width=1in,height=1.25in,clip,keepaspectratio]{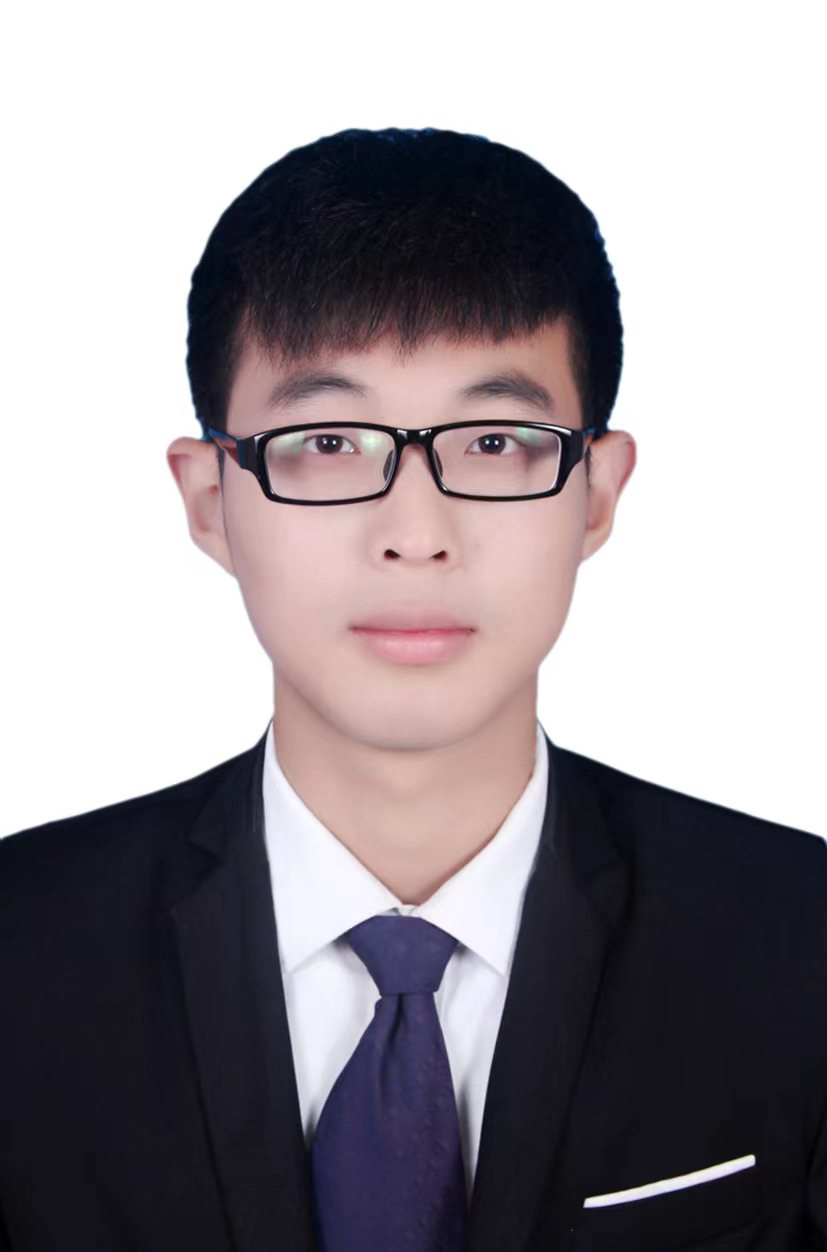}}]{Xiaoshuai Hao}
received the B.S. degree from Shandong University of Science and Technology China in 2017 and the Ph.D. degree from the Institute of Information Engineering, Chinese Academy of Sciences in 2023. 
He is currently a researcher in embodied multimodal large models at the Beijing Academy of Artificial Intelligence.
His research interests include multimedia retrieval, multimodal learning and embodied intelligence.
\end{IEEEbiography}
\vspace{-3 mm}
\begin{IEEEbiography}[{\includegraphics[width=1in,height=1.25in,clip,keepaspectratio]{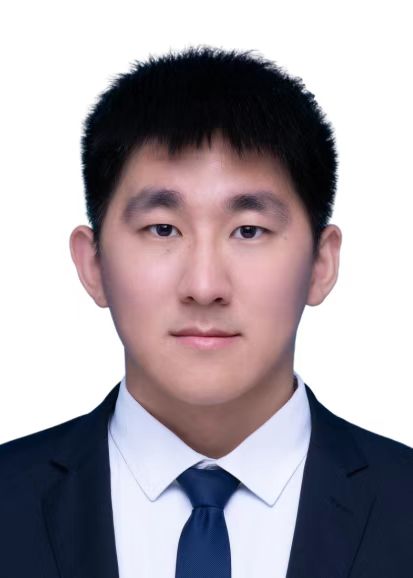}}]{Haichao Shi}
is an engineer at the Institute of Information Engineering, Chinese Academy of Sciences. He is a member of the Institute of Electrical and Electronics Engineers (IEEE), the International Computer Society (ACM), and the Chinese Society of Image and Graphics (CSIG). He has published more than twenty academic papers in TIP, TNNLS, PR, AAAI, MM, and other internationally recognized journals and conferences in related fields and has applied for five invention patents.
\end{IEEEbiography}
\vspace{-3 mm}
\begin{IEEEbiography}[{\includegraphics[width=1in,height=1.25in,clip,keepaspectratio]{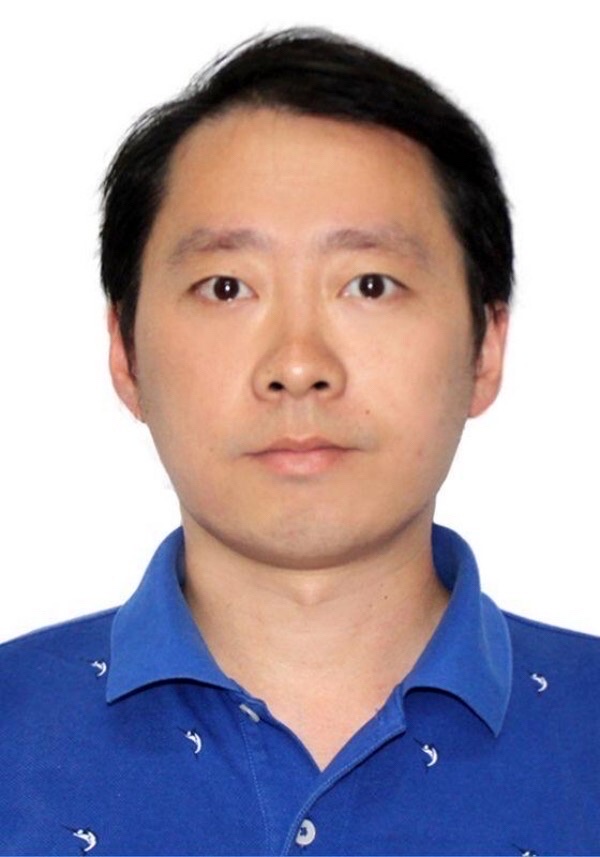}}]{Can Ma}
received his Ph.D. from the Institute of Computing Technology, Chinese Academy of Sciences, in 2012. He is currently a full senior engineer and Ph.D. supervisor at the Institute of Information Engineering, Chinese Academy of Sciences. His research interests include big data management and analysis in cyberspace and social networks.
\end{IEEEbiography}
\vspace{-3 mm}
\begin{IEEEbiography}[{\includegraphics[width=1in,height=1.25in,clip,keepaspectratio]{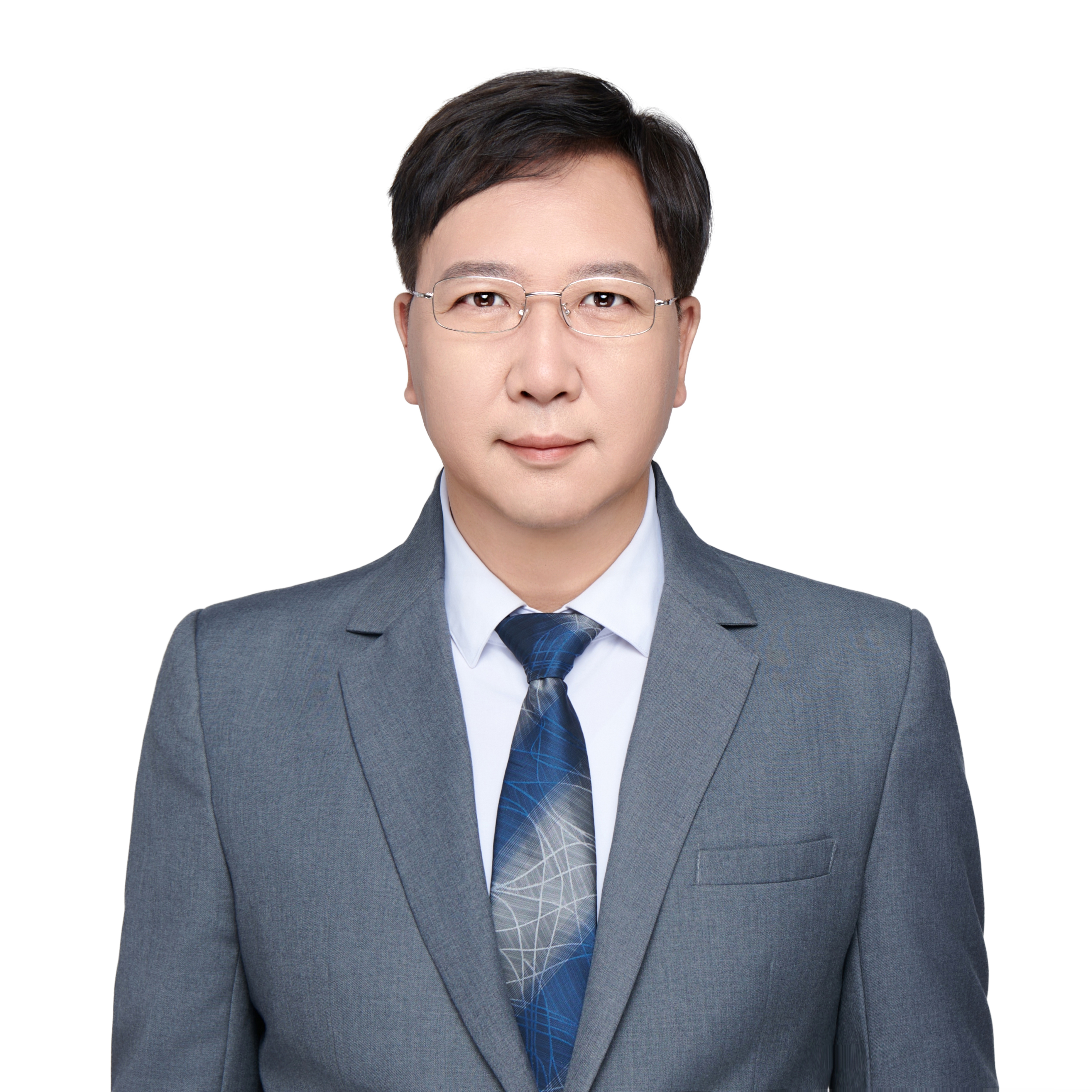}}]{Weiping Wang}
received a Ph.D. in Computer Science from the Harbin Institute of Technology, Harbin, China, in 2006. He is currently a Professor at the Institute of Information Engineering, Chinese Academy of Sciences, Beijing, China. His research interests include big data, data security, databases, and storage systems. He has more than 100 publications in major journals and international conferences.
\end{IEEEbiography}

\vfill

\end{document}